\documentclass[11pt]{article}
\DeclareUnicodeCharacter{2212}{\textminus}

\usepackage[final]{acl}
\usepackage{subcaption}
\usepackage{float}    % Cho phép dùng [H] để bảng không bị trôi
\usepackage{array} 
\usepackage{algorithm}
\usepackage{algpseudocode}
\usepackage{times}
\usepackage{latexsym}
\usepackage{CJKutf8}
\usepackage[T1]{fontenc}
\usepackage{booktabs}
\usepackage{multirow} 
\usepackage[utf8]{inputenc}
\usepackage{caption}
\usepackage{microtype}
\usepackage{enumitem} % Include enumitem package
\usepackage{listings}
\usepackage{algorithm}
\usepackage{float} % nếu dùng [H]
\usepackage{inconsolata}
\usepackage{amsmath} 
\usepackage{graphicx}
\usepackage{bm}
\usepackage{tikz}
\usetikzlibrary{bayesnet}
\usepackage{times}
\usepackage{blindtext}
\usepackage{microtype}
\usepackage{latexsym}
\usepackage{amsmath}
\usepackage{makecell}
\usepackage{amssymb}
\usepackage{dsfont}
\usepackage{graphicx}
\usepackage{subcaption}
\usepackage{multirow}
\usepackage{colortbl}
\usepackage[table]{xcolor}
\usepackage{booktabs}
\usepackage{threeparttable}
\usepackage{algorithm}
\usepackage[T1]{fontenc}

\usepackage[utf8]{inputenc}

\usepackage{microtype}

\usepackage{inconsolata}

\usepackage{graphicx}

\usepackage{multirow}
\usepackage[normalem]{ulem}
\useunder{\uline}{\ul}{}
\title{LLM-XTM: Enhancing Cross-Lingual Topic Models with Large Language Models}

\author{
  \textbf{Minh Chu Xuan\textsuperscript{1}\footnotemark[1]},
    \textbf{Tien-Phat Nguyen\textsuperscript{1}\footnotemark[1]}, \textbf{Dinh Viet Sang \textsuperscript{1}},\\
    \textbf{Linh Ngo Van\textsuperscript{1,\dag}},
  \textbf{Diep Thi-Ngoc Nguyen\textsuperscript{2}},
  \textbf{Trung Le\textsuperscript{3}}
  \bigskip \\
\textsuperscript{1}Hanoi University of Science and Technology, \\
\textsuperscript{2}VNU University of Engineering and Technology,
\textsuperscript{3}Monash University
}

\begin{document}
\maketitle
\renewcommand{\thefootnote}{\fnsymbol{footnote}}
\footnotetext[1]{Equal contribution}
\footnotetext[2]{Corresponding author: \href{mailto:email@domain}{ linhnv@soict.hust.edu.vn}}
\renewcommand*{\thefootnote}{\arabic{footnote}}
\begin{abstract} 
    
Cross-lingual topic modeling aims to discover shared semantic structures across languages, yet existing models depend on sparse bilingual resources and often yield incoherent or weakly aligned topics. Recent LLM-based refinements improve interpretability but are costly, document-level, and prone to hallucination, with prior white-box approaches requiring inaccessible token probabilities. We propose LLM-XTM, a framework that integrates LLM-guided topic refinement with self-consistency uncertainty quantification, enabling black-box, stable, and scalable enhancement of cross-lingual topic models. Experiments on multilingual corpora show that LLM-XTM achieves superior topic coherence and alignment while reducing reliance on bilingual dictionaries and expensive LLM calls. The code is publicly available at \url{https://github.com/tienphat140205/LLM-XTM}.

\end{abstract}

\section{Introduction}\label{section:introduction}

    Discovering latent themes within large text collections has long been a central problem in computational linguistics. Topic modeling (TM) provides the core framework for uncovering such hidden structures in an unsupervised manner~\cite{1999plsi, blei2003lda}. When extended to multilingual corpora, this paradigm becomes Cross-Lingual Topic Modeling (CLTM), which seeks to map semantically equivalent documents from different languages to comparable topic distributions~\cite{DBLP:conf/www/NiSHC09, MimnoWNSM09, YuanDY18, WuLZM20, infoctm, xtra, phat2026gloctm}. In addition to aligning topic proportions, CLTM must ensure that paired topics across languages convey consistent meanings, preserving interpretability and comparability across linguistic contexts. Consequently, CLTM serves as an essential framework for analyzing multilingual data and supporting cross-cultural semantic understanding.

    % Giảm padding mặc định giữa các cột
    \setlength{\tabcolsep}{1pt}     % Mặc định là 6pt
    
    % Giảm kích thước font bảng một chút
    \captionsetup[table]{font=small, skip=2pt}
    
    \begin{CJK}{UTF8}{gbsn}

    % Bảng đưa lên đầu cột và co lại theo chiều rộng cột
    \begin{table}[t]
    \centering
      % <-- Thêm dòng này
    \resizebox{\linewidth}{!}{%
    \begin{tabular}{llcccccc}
        \hline
        \textbf{En Topic\#1:} & & rating & gauge & height & mile & shoe  \\
        \textbf{ZH Topic\#1:} & & 投资者 & 财经 & 基金 & 股市 & 大盘  \\
        \textbf{Translation\#1:}   & & investor & finance & fund & stock mkt. & index  \\
        \hline
        \textbf{En Topic\#2:} & & buy & nice & awesome & excellent & seller  \\
        \textbf{ZH Topic\#2:} & & 好看 & 用品 & 说明书 & 精品 & 保养  \\
        \textbf{Translation\#2:}   & & good-looking & product & manual & premium & maintenance  \\
        \hline
        \textbf{En Topic\#3:} & & software & windows & desktop & tech & computer \\
        \textbf{ZH Topic\#3:} & & 体系 & 答 & 治疗 & 转换 & 整理 \\
        \textbf{Translation\#3:}   & & system & answer & treatment & convert & organize \\
        \hline
        \textbf{En Topic\#4:} & & news & hype & hint & reviews & media \\
        \textbf{ZH Topic\#4:} & & 炒作 & 皇帝 & 帅 & 想象 & 效应 \\
        \textbf{Translation\#4:}   & & hype & emperor & handsome & imagination & effect \\
        \hline
    \end{tabular}
    }
    \caption{Example of misaligned topics identified from the analysis. The words grouped under each topic differ in semantic coherence between the two languages by InfoCTM  \cite{infoctm}.}
    \label{tab:infoctm_misalignment}
    \vspace{-1em}  % Thu khoảng cách dưới bảng
    \end{table}

\end{CJK}
    
    Despite notable progress, most cross-lingual topic models still depend heavily on external bilingual resources—such as document embeddings, seed dictionaries, or parallel corpora—to establish alignment across languages \cite{MimnoWNSM09,infoctm, WuLZM20, refining, xtra}. While effective to some extent, these resources are inherently limited in coverage and often capture only surface-level or corpus-specific correspondences, providing shallow semantic signals that may not suffice to ensure deeper thematic consistency across languages \cite{VulicM15, YuanDY18}. In practice, the situation is further complicated by the quality of the corpora themselves: parallel or comparable datasets often contain mistranslations, domain mismatches, and lexical ambiguities, introducing substantial noise into the alignment process \cite{MimnoWNSM09, RuderVS19, ArtetxeS19}. Such imperfections make it difficult for models to discover truly coherent and semantically equivalent topics across languages. As a result, topics that are nominally aligned may diverge in meaning, and documents that express the same underlying content can still yield inconsistent topic distributions, as illustrated in Table ~\ref{tab:infoctm_misalignment}. Meanwhile, recent progress in large language models (LLMs) offers a promising remedy. Pretrained on massive multilingual corpora, LLMs internalize deep semantic correspondences that extend well beyond lexical overlap. Their ability to reason across languages makes them powerful semantic experts for refining topic-word distributions and strengthening cross-lingual alignment.

    While recent advances have explored the use of Large Language Models (LLMs) to refine or even directly generate topics, these approaches also face important limitations. Most existing methods treat LLM outputs as ground truth, prompting the model to produce topic words or document-level labels independently for each document \cite{Rijcken2023, Wang2023TopicLLM, Pham2024TopicGPT, Mu2024, Doi2024}. This design overlooks the global structure of the corpus, often leading to incomplete topic coverage and misalignment across documents. Moreover, running LLM inference for every document in large multilingual corpora is computationally prohibitive, making such approaches impractical at scale. Even when applied selectively, LLMs are prone to hallucination, producing irrelevant or misleading topic words that undermine consistency \cite{Ji2023HallucinationSurvey}. Attempts to address these issues, such as LLM-in-the-Loop (LLM-ITL) \cite{Yang2025LLMITL}, integrate LLM refinement during training, but rely on white-box access to token probabilities to estimate confidence. This requirement is costly and infeasible for low-resource settings or closed-source models. Consequently, there remains a need for a more efficient, uncertainty-aware way to incorporate LLM refinements into cross-lingual topic modeling.

    In this work, we propose LLM-XTM, a framework that integrates LLM refinement into cross-lingual topic modeling through a scalable, uncertainty-aware design. The first component focuses on refining topic-word distributions: candidate topic words produced by the base model are passed to an LLM for refinement, and multiple refinement paths are sampled. Inspired by SelfCheckGPT \cite{manakul2023selfcheckgpt}, which detects hallucinations in LLM outputs via self-consistency across generations, we adapt this technique to quantify uncertainty in topic refinement. Only words with high agreement across refinement paths are retained, producing stable and trustworthy refinements. These are then aligned with the base topic-word distributions using a distributional loss based on Maximum Mean Discrepancy (MMD) \cite{gretton2012kernel, li2015gmmn},ensuring that LLM feedback improves interpretability while remaining consistent with corpus-driven signals. This component enhances topic-word coherence and robustness, laying the foundation for reliable cross-lingual alignment.

    While LLM-guided topic-word refinement indirectly shapes document-topic distributions via the reconstruction loss, it does not explicitly enforce cross-lingual consistency at the document level. Due to lexical divergence, semantically equivalent documents in different languages may still yield diverging topic proportions $\theta_d$, as their surface word distributions differ significantly. To address this, we introduce a document-level alignment loss inspired by question answering: each document is encoded into a multilingual semantic space (e.g., using BGE-M3 \cite{bgem3}), and matched against topic embeddings derived from LLM-refined bilingual word sets, treating documents as “questions” and topics as “answers.” The resulting similarity-based topic distribution $\hat{\theta}_d$ reflects semantic relevance, and we minimize the KL-divergence $\text{KL}(\theta_d \parallel \hat{\theta}_d)$ to align latent topic proportions with semantic intent. This encourages semantically similar documents—regardless of language—to share similar topic distributions, thereby enforcing robust cross-lingual alignment at the document level.
Our contributions are threefold:
\begin{itemize}
    \item We introduce \textbf{LLM-XTM}, the first framework that enhances cross-lingual topic modeling with LLM-guided refinement while remaining efficient and scalable. 
    \item We adapt \textbf{self-consistency uncertainty quantification} to the topic refinement setting, enabling hallucination-resistant integration of LLM feedback, and align refined topic-word distributions with corpus-driven signals via an \textbf{MMD loss}.
    \item We propose a novel document-level alignment mechanism: refined topics are embedded into a multilingual space and matched with document embeddings, with a \textbf{KL-divergence objective} ensuring consistent alignment between semantic similarity vectors and $\theta$ distributions.
    \item Our extensive empirical results and qualitative analyses on diverse multilingual corpora not only show that LLM-XTM surpasses strong baselines but also provide insights into the effectiveness and robustness of each proposed component.
\end{itemize}

\section{Preliminaries} \label{sec:preliminaries}

\subsection{Notations} \label{subsec:notations}

 We model a multilingual corpus comprising $D$ documents in two languages, denoted $L_1$ and $L_2$, with the objective of discovering $K$ shared topics. The corpus is represented as a collection $X = \{x_d\}_{d=1}^D$ of Bag-of-Words (BoW) representations, where each document $x_d$ belongs to either $L_1$ or $L_2$. The vocabulary for $L_1$ is $V_1$ of size $|V_1|$, and for $L_2$ is $V_2$ of size $|V_2|$. The BoW representation of document $d$ is $x_d \in \mathbb{R}^{|V_\ell|}$, where $\ell \in \{1, 2\}$ indicates the document’s language. 
 
For each language $L_{\ell}$ where $\ell\in \{1, 2\}$, the topic-word distribution is denoted by $\beta^{(\ell)} \in \mathbb{R}^{|V_\ell| \times K} = (\beta_1^{(\ell)}, \ldots, \beta_K^{(\ell)})$, where each $\beta_k^{(\ell)} \in \mathbb{R}^{|V_\ell|}$ represents the word distribution for topic $k$ over the vocabulary $V_\ell$, satisfying $\displaystyle\sum_{v \in V_\ell} \beta_{v,k}^{(\ell)} = 1$. Each document $x_d$ is associated with a topic proportion vector $\theta_d \in \mathbb{R}^K$ such that $\displaystyle\sum_{k=1}^K \theta_{d,k} = 1$.

\begin{figure*}
     \centering
     \includegraphics[width=0.75\linewidth]{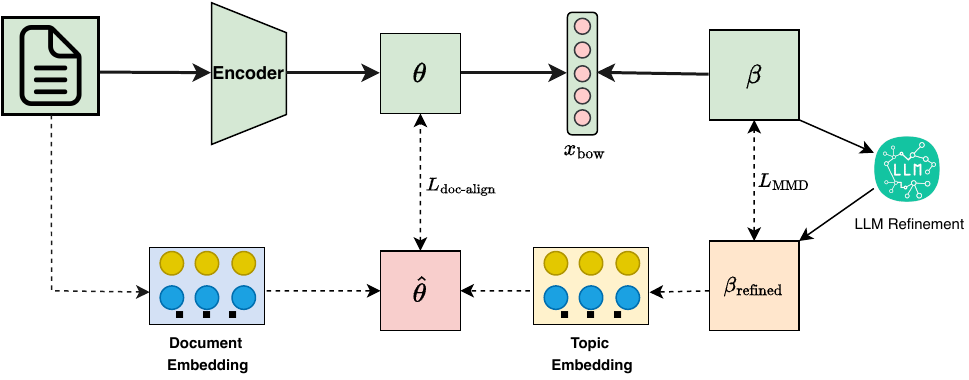}
     \caption{The LLM-XTM architecture enhances a VAE-based topic model using a dual-alignment strategy guided by a Large Language Model (LLM). The LLM first refines the model’s topic–word sets, producing semantically cleaner top-word lists ($\beta \rightarrow \beta_{\text{refined}}$). To ensure consistency between the original and refined topic representations, an MMD loss ($L_{\text{mmd}}$) encourages the original top-word distributions to stay close to the refined ones, preserving the latent semantic structure while enhancing coherence. Finally, a document-level alignment loss ($L_{\text{doc-align}}$) matches the model’s inferred document–topic distributions ($\theta$) with the LLM-induced semantic targets ($\hat{\theta}$), enforcing cross-lingual consistency across documents and topics.}
     \label{fig:model}
 \end{figure*} 
 
\subsection{VAE-based Topic Model} \label{subsec:VAE-based Topic Model}
We adopt a Variational Autoencoder (VAE) backbone. 
The encoder transforms each document’s Bag-of-Words vector $x_d$ into parameters $(\mu, \Sigma)$ of the posterior 
$q(z|x_d)=\mathcal{N}(z|\mu,\Sigma)$, from which a latent variable $z$ is sampled via reparameterization~\cite{kingma2013vae} 
under the Gaussian prior $p(z)=\mathcal{N}(z|\mu_0,\Sigma_0)$. 
Topic proportions are computed as $\theta_d=\mathrm{softmax}(z)$, 
and the decoder reconstructs $x_d$ from topic–word distributions $\beta \in \mathbb{R}^{V\times K}$ 
(optimized or embedding-based~\cite{srivastava2017prodlda,dieng2020etm,ecrtm}) using 
$\mathrm{Multinomial}(\mathrm{softmax}(\beta\theta_d))$. 
The training objective is:
\begin{equation*}
\begin{split}
\mathcal{L}_{\mathrm{TM}} = \frac{1}{D} \sum_{d=1}^{D} \Big[
    & -x_d^{\top} \log \mathrm{softmax}(\beta \theta_d) \\
    & + \mathrm{KL}(q(z|x_d)\|p(z))
\Big].
\end{split}
\end{equation*}
\subsection{Maximum Mean Discrepancy (MMD)} \label{subsec:mmd}
Let $k$ be a positive–definite kernel with RKHS $\mathcal{H}_k$. The kernel mean embedding maps a distribution $P$ to $\mu_k(P)=\mathbb{E}_{x\sim P}[k(x,\cdot)]\in\mathcal{H}_k$; with \emph{characteristic} kernels this map is injective, enabling metric comparison of distributions in RKHS \cite{muandet2017kernel,gretton2012kernel}. The squared Maximum Mean Discrepancy (MMD) between $P$ and $Q$ is
\begin{align}
\mathrm{MMD}_k^2(P,Q)
&= \mathbb{E}_{x,x'\sim P}\,k(x,x') \notag\\
&\quad + \mathbb{E}_{y,y'\sim Q}\,k(y,y') \notag\\
&\quad - 2\,\mathbb{E}_{x\sim P,\,y\sim Q}\,k(x,y).
\end{align}

which equals $\|\mu_k(P)-\mu_k(Q)\|_{\mathcal{H}_k}^2$ and serves as a principled two-sample discrepancy and alignment objective \cite{gretton2012kernel}. For weighted empirical samples $\{(x_i,w_i)\}$ and $\{(y_j,u_j)\}$ (weights sum to $1$), the quadratic estimator
$\sum_{i,i'} w_i w_{i'} k(x_i,x_{i'}) + \sum_{j,j'} u_j u_{j'} k(y_j,y_{j'}) - 2\sum_{i,j} w_i u_j k(x_i,y_j)$
is differentiable in both sample locations and weights, fitting our topic–word distributions. In embedding spaces we adopt a Gaussian kernel on cosine-induced distances and choose bandwidth via the median heuristic or a small multi-kernel set for scale robustness \cite{garreau2017median,muandet2017kernel}. MMD is closely connected to energy distance through distance-induced kernels, lending additional statistical justification \cite{sejdinovic2013equivalence}. We use this loss to align model and LLM-refined topic–word distributions (§\ref{sec:method}).

\section{Methodology}
\label{sec:method}
Our proposed framework, LLM-XTM, is a two-phase enhancement applied to a pre-trained cross-lingual topic model. Phase 1 corresponds to the backbone neural topic model (e.g., VAE-based InfoCTM, NMTM, or XTRA), which produces initial topic-word and document-topic distributions. The second phase, which is the focus of this section, applies the LLM-XTM enhancement architecture, illustrated in Figure~\ref{fig:model}, to refine and align the converged model. This enhancement stage consists of two primary components: (1) using a Large Language Model (LLM) to refine the topic-word distributions ($\beta$) for improved coherence, and (2) employing a novel question-answering (QA) inspired mechanism to align the document-topic distributions ($\theta$) for semantic consistency. The following sections will detail each component of this enhancement phase.

\subsection{Cross-Lingual Topic Word Refinement}

For each topic $k$, the base neural topic model yields two language-specific top-word lists 
$w^{(\text{en})}_k \subset V_{\text{en}}$ and $w^{(\text{zh})}_k \subset V_{\text{zh}}$ (top-15 each). 
We concatenate them into a bilingual candidate pool:
\[
C_k \;=\; w^{(\text{en})}_k \,\cup\, w^{(\text{zh})}_k.
\]

This pool is refined with a large language model (LLM)—for example, Gemini accessed via API, though any comparable LLM can be used.
The LLM is instructed to remove noisy or irrelevant items, 
retain the most representative words capturing the shared semantic theme, 
and add missing but coherent terms in both languages. 
The output is a fixed-size bilingual set $\bar{w}_k$ that improves 
interpretability and provides a stable basis for subsequent cross-lingual alignment.

To control the frequency of LLM interactions, we introduce a refinement frequency parameter f. The refinement process is executed every f epochs rather than at each training iteration, allowing the model to alternate between internal optimization and periodic LLM feedback. A smaller f yields more frequent refinements and tighter LLM guidance, whereas a larger f reduces computational overhead but updates less often (see \ref{subsec:param_sensitivity} for sensitivity analysis).

Implementation details and the exact prompt specification are provided in Appendix~\ref{app:prompt}.
\subsection{Self-Consistent Refinement}

While a single refinement pass with the LLM can improve topic word quality, the outputs remain inherently stochastic and may vary across runs. 
To enhance stability, we employ a self-consistent refinement strategy. 
For each topic $k$, the bilingual candidate pool $C_k$ is submitted to the LLM $R$ times, producing refined sets 
\[
\tilde{w}_k^{(1)}, \tilde{w}_k^{(2)}, \ldots, \tilde{w}_k^{(R)}.
\]

We then aggregate word occurrences across refinement rounds to estimate empirical frequencies:  
\[
f_k(v) \;=\; \frac{1}{R}\sum_{r=1}^R \mathbf{1}\{v \in \tilde{w}_k^{(r)}\}, \quad v \in \bigcup_{r=1}^R \tilde{w}_k^{(r)}.
\]

High-confidence words are selected as those with the largest $f_k(v)$, yielding the final refined bilingual set
\[
\bar{w}_k = \text{Top}_M \big( \{ (v, f_k(v)) \} \big).
\]

This self-consistency procedure filters out noisy candidates and preserves terms that consistently appear across multiple refinements, leading to a more reliable and interpretable bilingual representation of each topic. The number of refinement rounds R determines how many independent LLM refinements are aggregated, balancing stability and cost; its sensitivity is analyzed in Section \ref{subsec:param_sensitivity}.
\subsection{MMD-based Refinement Loss}
\label{sec:mmd}
To align the model's topic-word distributions with the LLM-refined sets, a loss based on the Maximum Mean Discrepancy (MMD) is used \cite{gretton2012kernel,muandet2017kernel}. This approach requires creating two distinct distributions for each topic \(k\): the model's raw distribution and the LLM's refined target distribution.

The model's raw distribution, \(\beta_{k}^{(raw)}\), is constructed from the top-N words generated by the decoder. The probabilities for the top words in each language are combined into a language-balanced mixture, which is then normalized as follows:
\[
\tilde{\beta}_{k}^{(raw)}(v) = 
\begin{cases}
    \beta_{k}^{(l1)}(v), & v \in W_{k}^{(l1)} \\
    \beta_{k}^{(l2)}(v), & v \in W_{k}^{(l2)} \\
    0, & \text{otherwise}
\end{cases}
\]
\[
\beta_{k}^{(raw)}(v) = \frac{\tilde{\beta}_{k}^{(raw)}(v)}{\sum_{u\in U_{k}}\tilde{\beta}_{k}^{(raw)}(u)}
\]
where \(W_{k}^{(l)}\) is the top-word list for language \(l\), and \(U_k\) is the union of these lists.

The LLM-refined target distribution, \(\beta_{k}^{(refined)}\), is derived from the self-consistent refinement process where word counts across multiple refinement rounds are tallied. These counts are similarly formed into a language-balanced mixture and normalized to create the target:
\[
\tilde{c}_{k}(v) = 
\begin{cases}
    c_{k}^{(l1)}(v), & v \in \overline{W}_{k}^{(l1)} \\
    c_{k}^{(l2)}(v), & v \in \overline{W}_{k}^{(l2)} \\
    0, & \text{otherwise}
\end{cases}
\]
\[
\beta_{k}^{(refined)}(v) = \frac{\tilde{c}_{k}(v)}{\sum_{u\in\tilde{U}_{k}}\tilde{c}_{k}(u)}
\]
where \(\overline{W}_{k}^{(l)}\) is the refined word set for language \(l\), and \(\overline{U}_k\) is the union of these sets.

The divergence between these two distributions is measured using the squared MMD, calculated with a Gaussian kernel on the cosine distances between word embeddings \cite{garreau2017median,sejdinovic2013equivalence}. The final training objective is the average MMD loss over all \(K\) topics. This loss shifts the model's topic-word distributions toward the more coherent LLM-refined targets.
\begin{align*}
\mathcal{L}_{\text{MMD}} = \frac{1}{K} \sum_{k=1}^{K} \text{MMD}^{2}(\beta_{k}^{\text{(raw)}}, \beta_{k}^{\text{(refined)}})
\label{mmd_loss}
\end{align*}

\subsection{Document--Topic Alignment via QA Mechanism}
\label{sec:doc_align}
While refining topic words enhances coherence, a core challenge remains: ensuring that documents with the same meaning in different languages are assigned similar topic distributions. To solve this, we introduce a pioneering document-level alignment mechanism that reframes the problem through the intuitive lens of Question Answering (QA).

In this elegant paradigm, each document is treated as a "question" asking, "What is my core semantic theme?". The LLM-refined topics, in turn, act as a pool of high-quality "candidate answers". The goal is to teach the model how to match each question to its most relevant answer, irrespective of the document's language.

The mechanism works by first transforming each LLM-refined bilingual topic word list ($\overline{w}_{k}$) into a rich semantic vector using a powerful multilingual encoder like BGE-M3 \cite{bgem3}. This creates a single, potent embedding for each topic ($t_{k} = \text{Enc}(\overline{w}_{k})$), representing the core meaning of the "answer". In parallel, every document ($x_{d}$) is mapped into the same semantic space, producing a corresponding "question" vector ($h_{d} = \text{Enc}(x_{d})$). From here, we calculate the cosine similarity between each document "question" ($h_d$) and all topic "answers" ($\{t_{k}\}_{k=1}^{K}$):  
\[
s_{d,k} = \frac{h_d^\top t_k}{\|h_d\|_2 \, \|t_k\|_2}, \qquad k=1,\dots,K
\]  
These similarity scores are normalized via a softmax to obtain a new topic distribution that purely reflects semantic relevance:  
\[
\hat{\theta}_{d,k} = \frac{\exp(s_{d,k}/\tau)}{\sum_{j=1}^{K}\exp(s_{d,j}/\tau)}.
\]

Finally, we align the model's internal topic posterior ($\theta_{d}$) with this powerful external semantic signal by minimizing the KL-divergence between the two distributions:  
\[
L_{\text{doc-align}}=\sum_{d=1}^{D}KL(\theta_{d}||\hat{\theta}_{d}).
\]  
This objective forces the model to learn cross-lingually consistent document representations, ensuring that semantic meaning, not just vocabulary, drives the topic assignments.

\begin{table*}[!htbp]
\centering
\small
% giảm khoảng cách cột để tiết kiệm chiều ngang
\setlength{\tabcolsep}{2.2mm}
\renewcommand{\arraystretch}{1.1}

% bó bảng theo đúng bề rộng hai cột
\resizebox{\textwidth}{!}{%
\begin{tabular}{l ccc ccc ccc}
\toprule
\multirow{2}{*}{\textbf{Model}} 
  & \multicolumn{3}{c}{\textbf{EC News}} 
  & \multicolumn{3}{c}{\shortstack{\textbf{Amazon Review}}} 
  & \multicolumn{3}{c}{\shortstack{\textbf{Rakuten Amazon}}} \\
\cmidrule(lr){2-4} \cmidrule(lr){5-7} \cmidrule(lr){8-10}
 & CNPMI & TU & TQ & CNPMI & TU & TQ & CNPMI & TU & TQ \\
\midrule
MCTA
& 0.025 & 0.489 & 0.012 & 0.028 & 0.319 & 0.009 & 0.021 & 0.272 & 0.006 \\

MTAnchor
& $-0.013$ & 0.192 & 0.000 & 0.028 & 0.323 & 0.009 & $-0.001$ & 0.214 & 0.000 \\

u\text{-}SVD
& 0.082 & 0.830 & 0.068 & 0.055 & 0.634 & 0.035 & 0.027 & 0.571 & 0.015\\

SVD-LR
& 0.083 & 0.820 & 0.068 & 0.053 & 0.627 & 0.033 & 0.026 & 0.558 & 0.015\\
\midrule
XTRA 
& 0.078 & 0.978 & 0.076 
& 0.053 & 0.979 & 0.052 
& 0.034 & 0.966 & 0.033 \\
+ LLM-XTM 
& 0.088 & 0.954 & 0.084 
& 0.072 & 0.959 & 0.069 
& 0.037 & 0.945 & 0.035 \\

& $\uparrow$12.8\% & $\downarrow$2.5\% & $\uparrow$10.5\%
& $\uparrow$35.8\% & $\downarrow$2.0\% & $\uparrow$32.7\%
& $\uparrow$8.8\% & $\downarrow$2.2\% & $\uparrow$6.1\% \\
\midrule
InfoCTM 
& 0.041 & 0.943 & 0.039 & 0.037 & 0.930 & 0.034 & 0.032 & 0.870 & 0.028
\\
{+ LLM-XTM} 
& 0.062 & 0.898 & 0.056 & 0.050 & 0.933 & 0.047 & 0.040 & 0.870 & 0.035 \\

& $\uparrow$\,51.2\% & $\downarrow$\,4.8\% & $\uparrow$\,43.6\%
& $\uparrow$\,35.1\% & $\uparrow$\,0.3\% & $\uparrow$\,38.2\%
& $\uparrow$\,25.0\% & $\uparrow$\,0.0\% & $\uparrow$\,25.0\% \\
\midrule
NMTM 
& 0.034 & 0.818 & 0.028 & 0.043 & 0.610 & 0.026 & 0.012 & 0.633 & 0.008
 \\
{+ LLM-XTM} 
& 0.039 & 0.821 & 0.032 & 0.056 & 0.627 & 0.035 & 0.016 & 0.666 & 0.011
\\

& $\uparrow$\,15.9\% & $\uparrow$\,0.4\% & $\uparrow$\,14.3\%
& $\uparrow$\,30.2\% & $\uparrow$\,2.8\% & $\uparrow$\,34.6\%
& $\uparrow$\,33.3\% & $\uparrow$\,5.2\% & $\uparrow$\,37.5\% \\

\bottomrule
\end{tabular}%
}
\caption{Topic coherence (CNPMI), topic uniqueness (TU), and topic quality (TQ) where TQ = max(0, CNPMI) × TU. Results for MCTA and MTAnchor are taken directly from \cite{infoctm}, while all other baselines and our model are tuned and averaged over five random seeds, with the mean values reported.}

\label{tab:topic_quality}
\end{table*}

\subsection{Overall Objective} \label{subsec:overall-objective}

Our framework operates in two phases. 
\textbf{Phase~1} trains a base neural topic model (e.g., NMTM, InfoCTM), 
optimized with its original objective, which may consist of $\mathcal{L}_{\mathrm{TM}}$ 
and additional backbone-specific losses (e.g., contrastive or clustering terms). 
\textbf{Phase~2} then enhances the converged model by incorporating our refinement 
and alignment components. The composite objective in Phase~2 over encoder parameters 
$\phi$ and decoder parameters $\psi$ is:
\begin{align*}
\label{eq:overall}
\mathcal{J}(\phi,\psi) 
&= \mathcal{L}_{\text{Phase~1}} 
+ \lambda_{\mathrm{mmd}}\,\mathcal{L}_{\mathrm{MMD}} \notag\\
&\quad + \lambda_{\mathrm{qa}}\,\mathcal{L}_{\mathrm{doc-align}}.
\end{align*}
where $\mathcal{L}_{\text{Phase1}}$ denotes the loss function used by the chosen 
backbone model in Phase~1, while $\mathcal{L}_{\mathrm{MMD}}$ (Section~\ref{sec:mmd}) 
and $\mathcal{L}_{\mathrm{doc-align}}$ (Section~\ref{sec:doc_align}) are the additional 
enhancement losses applied in Phase~2.

\section{Experiments and Results}

\subsubsection*{Datasets}
We evaluated on three public benchmarks: \textbf{EC News}~\cite{WuLZM20}, a bilingual English–Chinese news corpus spanning six domains; \textbf{Amazon Review}~\cite{YuanDY18}, English–Chinese product reviews converted to a binary task (five-star = 1, others = 0); and \textbf{Rakuten Amazon}~\cite{YuanDY18}, Japanese–English reviews similarly framed as binary classification by rating. Detailed dataset statistics are provided in \ref{sec:dataset_statistics}

\subsubsection*{Baseline Models}
We compared our method with a range of representative cross-lingual topic modeling baselines. \textbf{MCTA}~\cite{DBLP:conf/acl/ShiLBX16} is a probabilistic CLTM framework that models cultural variation across languages. \textbf{MTAnchor}~\cite{YuanDY18} aligns topics through multilingual anchor words. \textbf{NMTM}~\cite{WuLZM20} introduces a neural approach that embeds multilingual topics in a shared latent space. \textbf{InfoCTM}~\cite{infoctm} enhances topic representations using mutual information maximization to reduce redundancy. \textbf{XTRA}~\cite{xtra} extends this line by applying contrastive learning jointly on document–topic ($\theta$) and topic–word ($\beta$) spaces to achieve dual alignment. Finally, we include two clustering-based refinement methods—\textbf{u-SVD} and \textbf{SVD-LR}~\cite{refining}—as modern baselines for topic post-processing and coherence enhancement.
\subsubsection*{Evaluation Metrics}
We assess topic quality and utility through both intrinsic and extrinsic evaluations. Intrinsically, \textbf{CNPMI}~\cite{cnpmi} measures cross-lingual coherence, \textbf{TU}~\cite{tu} quantifies topic diversity using the top 15 words, and \textbf{TQ}~\cite{refining} combines both with negative coherence clipped to zero. We further include \textbf{LLM-based ratings}~\cite{llm_eval} (1–3 scale) to evaluate intra-lingual coherence and cross-lingual alignment.

\begin{table*}[t]
\centering
\small
\setlength{\tabcolsep}{3pt}
\begin{tabular}{l c c c c c c}
\toprule
& \multicolumn{2}{c}{Topic Quality} & \multicolumn{4}{c}{Classification} \\
\cmidrule(lr){2-3} \cmidrule(lr){4-7}
Model                          & CNPMI & TU    & EN-I  & JA-I  & EN-C  & JA-C  \\
\midrule
NMTM$^{\dagger}$              & 0.012 & 0.633 & 0.796 & 0.826 & 0.610 & 0.681 \\ 
NMTM + LLM-XTM$^{\dagger}$           & 0.016 & 0.666 & 0.792 & 0.833 & 0.621 & 0.728 \\
\midrule
\textbf{NMTM + LLM-XTM (w/o $L_{\text{doc-align}}$)} & 0.012 & 0.679 &0.791 & 0.832 & 0.611 & 0.723 \\ 
\textbf{NMTM + LLM-XTM (w/o $L_{\text{MMD}}$)}       & 0.012 & 0.641 & 0.795 & 0.833 & 0.621 & 0.723 \\
\textbf{NMTM + LLM-XTM (w/o self-consistency)}       & 0.011 & 0.654 & 0.792 & 0.832 & 0.619 & 0.720 \\
\textbf{NMTM + LLM-XTM (w/ $L_{\text{OT}}$)}       & 0.013 & 0.664 & 0.795 & 0.829 & 0.620 & 0.720 \\
\midrule
\bottomrule
\end{tabular}
\caption{Ablation study on Rakuten Amazon (50 topics). We report Topic Quality (CNPMI, TU) and Classification Accuracy. Baselines ($\dagger$) are reported from \cite{infoctm}. Ablations are applied to the LLM-XTM framework on the \emph{NMTM} backbone.}
\label{tab:ablation_ecnews}
\end{table*}

\subsection{Topic Quality Analysis}  

Table~\ref{tab:topic_quality} presents intrinsic metrics for topic coherence (CNPMI), diversity (TU), and composite topic quality (TQ) across three benchmark datasets. Integrating \textbf{LLM-XTM} with \textbf{XTRA} consistently improves topic quality: on EC News, CNPMI increases by 12.8\% and TQ by 10.5\%; on Amazon Review, coherence improves by 35.8\% and TQ by 32.7\%. The slight reduction in TU (\(\approx 2\%\)) suggests that redundant or peripheral words are pruned, yielding more semantically focused topics. On Rakuten Amazon, CNPMI rises by 8.8\% and TQ by 6.1\%, indicating that LLM-based refinement enhances cross-lingual consistency while maintaining reasonable topic diversity.

Across other backbones, \textbf{LLM-XTM} shows even larger impacts. For \textbf{InfoCTM}, CNPMI jumps 51.2 \% on EC News and 35.1 \% on Amazon Review, with corresponding TQ gains of 43.6 \% and 38.2 \%; TU decreases are small (under 5 \%). On the lighter \textbf{NMTM} model, LLM-XTM consistently boosts coherence and quality, with CNPMI gains in the range 15.9–30.2 \% and TQ improvements up to 37.5 \%. These findings confirm that LLM-XTM acts as a general enhancement layer, reliably increasing topic coherence and cross-lingual interpretability while exerting only minimal negative impact on topic diversity.

\medskip  
\noindent\textbf{On the slight TU reduction.}  
Note that TU (topic uniqueness) is usually defined as the average reciprocal of the number of occurrences of each top word across all topics (i.e.\ if a word appears in multiple topics, its contribution to TU is reduced). Hence, even small overlaps in high-importance words across topics decrease TU, regardless of whether those overlaps are semantically justified. In our setting, LLM-XTM favors core semantic terms and removes noisy or weakly related terms, so multiple topics may retain some shared “meaningful” words. This overlap leads to a modest TU drop despite improvements in coherence. The TU decline thus should be interpreted not purely as loss of diversity, but as a side-effect of focusing topics toward more semantically consistent vocabularies.  

\subsection{Evaluation of Topic Distributions via Classification}

\subsection{Ablation Study} \label{subsec:ablation}
Table~\ref{tab:ablation_ecnews} reports an ablation on Rakuten Amazon with 50 topics to analyze each component’s effect. The full \textbf{LLM-XTM} improves topic quality over NMTM$^{\dagger}$, reaching CNPMI 0.016 and TU 0.666 (vs. 0.012/0.633), and also enhances classification performance (EN-C 0.621, JA-C 0.728). Removing \(L_{\text{doc-align}}\) notably weakens cross-lingual alignment, reducing CNPMI to 0.012 and EN-C to 0.611. Excluding \(L_{\text{MMD}}\) impairs topic–word coherence, dropping TU to 0.641 and JA-C to 0.723, while omitting self-consistency further degrades both CNPMI (0.011) and TU (0.654) with minor accuracy losses. Finally, comparing discrepancy measures, MMD surpasses Optimal Transport (CNPMI 0.016 vs. 0.013), as its kernel-based design more effectively captures semantic relations in shared multilingual embedding space, making it better suited for cross-lingual topic modeling.

\subsection{LLM-based Topic Quality Evaluation}

\begin{figure}[t]
    \centering
    % Subfigure 1
    \begin{subfigure}[b]{1\linewidth}
        \centering
        \includegraphics[width=\linewidth]{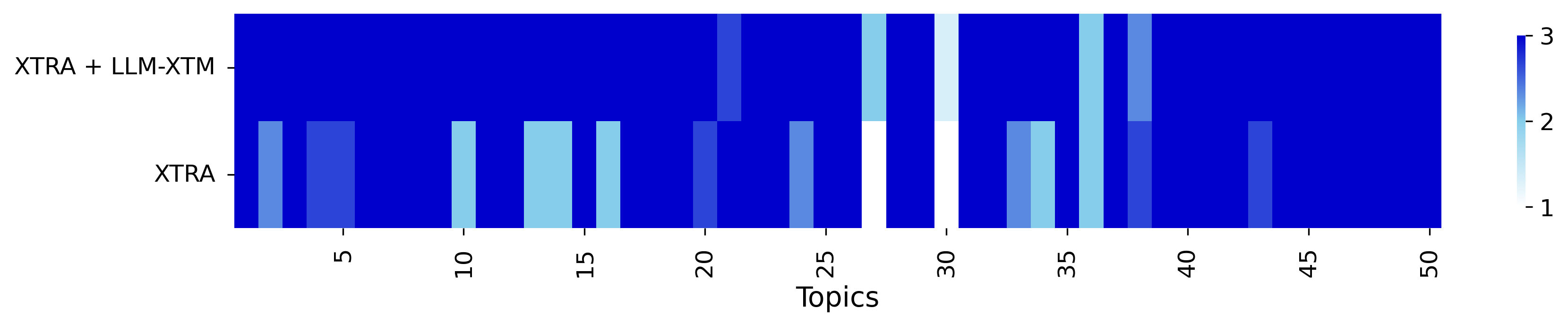}
        \caption{English intra-lingual semantic similarity}
        \label{fig:A1}
    \end{subfigure}
    \vspace{0.5em}

    % Subfigure 2
    \begin{subfigure}[b]{1\linewidth}
        \centering
        \includegraphics[width=\linewidth]{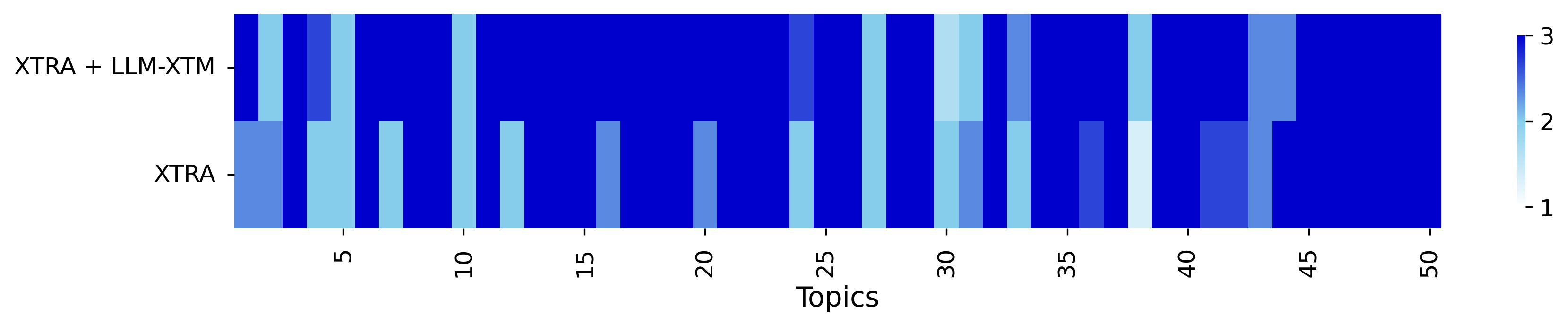}
        \caption{Chinese intra-lingual semantic similarity}
        \label{fig:A2}
    \end{subfigure}
    \vspace{0.5em}

    % Subfigure 3
    \begin{subfigure}[b]{1\linewidth}
        \centering
        \includegraphics[width=\linewidth]{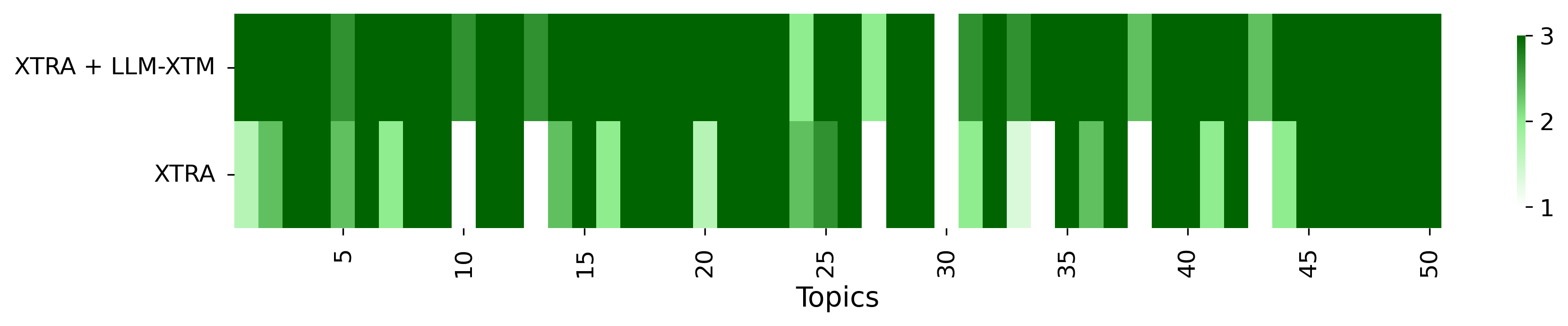}
        \caption{Cross-lingual semantic similarity on Amazon Review dataset.}
        \label{fig:A}
    \end{subfigure}
    \caption{LLM-based evaluations of inner and cross-lingual semantic similarity on dataset Amazon Review. Darker shades indicate higher scores (on a 1–3 scale). Reported values are rounded from the mean scores over four independent LLM runs.}

    \label{fig:llm_eval}
\end{figure}

We conducted an automated evaluation using a Large Language Model (LLM) to quantitatively assess the impact of \textbf{LLM-XTM} on the strong \textbf{XTRA} baseline, following recent topic-model evaluation practices~\cite{llm_eval}. As shown in Figure~\ref{fig:llm_eval}, the LLM rated intra-lingual coherence and cross-lingual alignment on a 1--3 scale. Results show consistent improvements in both aspects, with notably stronger cross-lingual alignment where the baseline was unstable. Overall, \textbf{LLM-XTM} markedly enhances semantic consistency and alignment across languages, confirming its effectiveness as a refinement framework. The full evaluation script is provided in Appendix~\ref{app:prompt}, and additional results are detailed in Appendix~\ref{sec:llm_results}.

\subsection{Qualitative Analysis: Discovered Topic Word Examples}
\label{sec:qualitative_topic_analysis_appendix}

\begin{CJK}{UTF8}{gbsn}
Table~\ref{tab:topic_examples_music_fashion_care} illustrates cross-lingual topic examples in \textit{Music \& Singing}, \textit{Fashion \& Magazines}, and \textit{Personal Care / Cosmetics}, comparing baseline models with their LLM-enhanced versions (NMTM + LLM-XTM, InfoCTM + LLM-XTM, and XTRA + LLM-XTM). Upper blocks show noisy or misaligned keywords, while lower blocks present refined, coherent sets. Baseline models often include off-topic terms such as \textcolor{red}{``嫁''} (marry), \textcolor{red}{``上网''} (go online), or \textcolor{red}{``旅行''} (travel), which blur topical focus and weaken bilingual alignment. In contrast, LLM-XTM removes noise and strengthens cross-lingual coherence—for instance, in \textit{Cosmetics}, (\textit{skin}, \textit{fragrance})~$\leftrightarrow$~(\textit{保湿}, \textit{洗}) form clean, semantically aligned pairs—demonstrating clear improvements in both coherence and semantic consistency across languages.
\end{CJK}

\begin{CJK}{UTF8}{gbsn}
\begin{table}[!h]
\centering
\resizebox{\linewidth}{!}{%
\begin{tabular}{@{}llccccc@{}}
    \toprule
    \multicolumn{7}{c}{\textbf{Topic: Music \& Singing \quad (NMTM — Amazon Review, Topic 26)}} \\
    \midrule
    \multicolumn{7}{c}{NMTM, Topic 26 (Poor)} \\
    \midrule
    \textbf{EN:} & & song & albums & lyrics & riffs & catchy \\
    \textbf{ZH:} & & \textcolor{red}{思念} & \textcolor{red}{挚爱} & \textcolor{red}{誓言} & \textcolor{red}{嫁} & \textcolor{red}{信件} \\
    \textbf{Translations:} & & longing & true love & vow & marry & letter \\
    \midrule
    \multicolumn{7}{c}{NMTM + LLM-XTM, Topic 26 (Good)} \\
    \midrule
    \textbf{EN:} & & vocals & singer & lyrics & album & drums \\
    \textbf{ZH:} & & 音乐 & 专辑 & 歌手 & 歌迷 & 浪漫 \\
    \textbf{Translations:} & & music & album & singer & fan & romantic \\
    \midrule\midrule

    \multicolumn{7}{c}{\textbf{Topic: Fashion \& Magazines \quad (InfoCTM — ECNews, Topic 45)}} \\
    \midrule
    \multicolumn{7}{c}{InfoCTM, Topic 45 (Poor)} \\
    \midrule
    \textbf{EN:} & & cover & magazine & campaign & bundchen & \textcolor{red}{fail} \\
    \textbf{ZH:} & & \textcolor{red}{外壳} & \textcolor{red}{上网} & 采访 & 付费 & 封面 \\
    \textbf{Translations:} & & shell/casing & go online & interview & pay & cover \\
    \midrule
    \multicolumn{7}{c}{InfoCTM + LLM-XTM, Topic 45 (Good)} \\
    \midrule
    \textbf{EN:} & & vogue & cover & supermodel & magazine & shoot \\
    \textbf{ZH:} & & 封面 & 模特 & 摄影 & 杂志 & 时尚 \\
    \textbf{Translations:} & & cover & model & photography & magazine & fashion \\
    \midrule\midrule

    \multicolumn{7}{c}{\textbf{Topic: Personal Care / Cosmetics \quad (XTRA — Rakuten Amazon, Topic 40)}} \\
    \midrule
    \multicolumn{7}{c}{XTRA, Topic 40 (Poor)} \\
    \midrule
    \textbf{EN:} & & makeup & \textcolor{red}{festival} & hair & lipstick & \textcolor{red}{camera} \\
\textbf{JA:} & & 美容 & \textcolor{red}{旅行} & 皮肤 & 化粧 & 保养 \\
\textbf{Translations:} & & beauty\ makeup & travel & skin & cosmetics & skincare \\
    \midrule
    \multicolumn{7}{c}{XTRA + LLM-XTM, Topic 40 (Good)} \\
    \midrule
    \textbf{EN:} & & skin & shampoo & fragrance & sensitive & lotion \\
    \textbf{JA:} & & 化粧 & 保湿 & 洗顔 & 頭皮 & 敏感 \\
    \textbf{Translations:} & & makeup & moisturizing & face wash & scalp & sensitive \\
    \bottomrule
\end{tabular}%
}
\caption{Cross-lingual topic examples. Upper blocks show noisy/misaligned sets; lower blocks show coherent sets. \textcolor{red}{Red} marks tokens that are semantically off-topic or misaligned.}
\label{tab:topic_examples_music_fashion_care}
\end{table}
\end{CJK}

\subsection{Sensitivity Analysis of Refinement Parameters} 
\label{subsec:param_sensitivity} 
We study two hyperparameters of our LLM-based refinement: refinement rounds ($R$) and refinement frequency ($f$, epochs between LLM calls). Experiments use the Amazon Review dataset with NMTM, holding other settings fixed. We test $R \in \{1,3,5,7,10,13\}$ and $f \in \{5,8,10,13\}$. 
\begin{figure}[!htbp] 
    \centering 
    \includegraphics[width=0.48\textwidth]{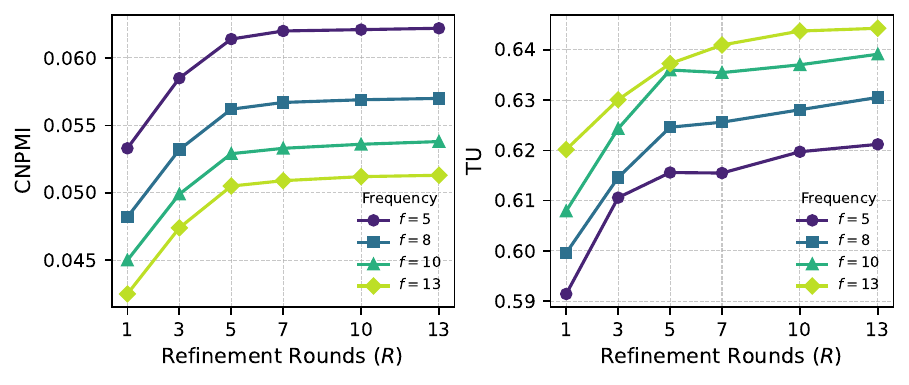} 
    \caption{Sensitivity to rounds ($R$) and frequency ($f$) in CNPMI (left) and TU (right) on Amazon Review.} 
    \label{fig:sensitivity_combined} 
\end{figure} 
\textbf{Refinement rounds ($R$).} 
Aggregating $R$ independent LLM calls improves CNPMI while TU shows mixed behavior. At $f=8$, CNPMI increases from 0.0482→0.0562 (+16.6\%) as $R$ rises from 1 to 5, but TU rises only marginally from 0.600→0.627 (+4.2\%). Beyond $R=5$, gains diminish: from $R=5$ to $R=13$, CNPMI improves just 1.4\% (0.0562→0.0570) while TU increases 1.0\% (0.625→0.631). Since cost scales linearly with $R$, settings around $R=5$–7 capture most benefits efficiently.

\textbf{Refinement frequency ($f$).} 
Lower $f$ (more frequent calls) boosts CNPMI but reduces TU. At $R=5$, CNPMI drops 17.8\% (0.614→0.505) while TU increases 3.5\% (0.616→0.637) when $f$ increases from 5 to 13, revealing a clear trade-off between coherence and uniqueness.

\section{Conclusion} \label{sec:conclusion}
We introduced \textbf{LLM-XTM}, a two-stage framework that applies a post-hoc enhancement to cross-lingual topic models: (i) we refine topic–word distributions with an LLM using a self-consistency–based filter and an MMD alignment loss, and (ii) we align document–topic posteriors to semantic targets via a QA-style objective. It improves cross-lingual coherence and transfer while preserving diversity and serves as a plug-in for diverse topic model backbones.
\section*{Limitation}
The efficacy of LLM-XTM is fundamentally tethered to the quality of the underlying topic model it enhances. While our framework significantly improves coherence and alignment, its performance ceiling is constrained by the initial topics provided by the backbone model; it can refine, but not create, sensible topics from a completely failed initialization. Moreover, the reliance on multiple API calls to external LLMs for self-consistent refinement introduces considerations of computational latency, making the enhancement phase less suitable for real-time or highly resource-constrained environments.


\section*{Ethical Considerations}
We adhere to the ACL Code of Ethics and the terms
of each codebase license. Our method aims to
advance the field of topic modeling, and we are
confident that, when used properly and with care,
it poses no significant social risks.

\section*{Acknowledgments}
Trung Le was supported by the Air Force Office of Scientific Research under award number FA2386-25-1-4023 and the ARC Discovery Project grant DP250100262.

\bibliography{custom}

\appendix

\newpage
\section{Related Work}

\textbf{Topic Models and Cross-Lingual Extensions.}
Topic modeling seeks to identify hidden semantic patterns within large text collections. The foundational Latent Dirichlet Allocation (LDA)~\cite{blei2003lda} conceptualizes documents as mixtures of latent topics, establishing the basis for later developments. Neural Topic Models (NTMs) build upon this idea through deep generative formulations such as variational autoencoder (VAE) frameworks, including NVDM and ProdLDA~\cite{srivastava2017prodlda}, as well as embedding-oriented approaches like ETM~\cite{dieng2020etm}. Subsequent research integrates pre-trained language models (e.g., BERT) to better capture contextual and semantic nuances~\cite{BianchiTH20,BianchiTHNF21,HoyleGR20}. Other lines of work leverage optimal transport techniques~\cite{nstm,ecrtm} or contrastive learning objectives~\cite{2021contrastiventm,NguyenWDNNL24}, with more recent approaches combining the two paradigms~\cite{cite4,chi2026ot}, as well as mutual-information-based regularizers~\cite{pham2024neuromaxenhancingneuraltopic}, to improve topic coherence and representation separation. Complementary efforts target the training procedure itself, ranging from sharpness-aware minimization~\cite{cite6} to multi-objective surrogate formulations~\cite{chi2026sur,tue2025moo}. In parallel, clustering-based approaches such as Top2Vec~\cite{top2vec} and BERTopic~\cite{grootendorst2022bertopic} derive topics directly from dense embeddings, enhancing interpretability and flexibility across datasets, while a growing body of work tackles short-text, sub-document, or temporal settings where BoW signals are sparse or drift over time~\cite{cite5,cite3,tung2026global,nguyen2025sub,nguyen2026beyond}. Overall, research in this area has shifted from traditional count-based probabilistic models toward representation-driven frameworks that emphasize semantic alignment and contextual understanding. 

Cross-lingual topic modeling (CLTM) extends topic modeling to multilingual settings to discover aligned themes across languages. Early work such as \citet{MimnoWNSM09} relied on parallel corpora, limiting broader applicability. Later methods leveraged bilingual dictionaries to align vocabularies \cite{JagarlamudiD10,abs-1205-2657}, with translation-based refinements proposed by \citet{DBLP:conf/acl/ShiLBX16}, \citet{YuanDY18}, \citet{YangBR19}, \citet{WuLZM20}, and \citet{infoctm}. A more recent line, XTRA~\cite{xtra}, applies contrastive learning simultaneously on $\theta$ (document--topic distributions) and $\beta$ (topic--word projections) to achieve dual semantic alignment; concurrently, GloCTM~\cite{phat2026gloctm} pursues cross-lingual alignment through a shared global context space rather than relying on bilingual lexicons. Another strand employs multilingual embeddings~\cite{ChangH21}, but these often assume isomorphic embedding spaces that may not hold for distant languages. Transformer-based and zero-shot models (e.g., \cite{BianchiTHNF21, MuellerD21}) reduce dependence on parallel data yet still struggle with consistent cross-lingual alignment. In parallel, refinement methods for clustering-based CLTM, such as u-SVD and SVD-LR~\cite{refining}, remove language-dependent dimensions from embeddings before clustering, improving alignment robustness.

Furthermore, as cross-lingual topic modeling increasingly relies on dense representations, embedding alignment and distillation have emerged as critical areas of research. In our QA-inspired alignment mechanism, we rely on powerful pre-trained multilingual encoders like BGE-M3 \cite{bgem3} to establish a shared semantic space. However, recent advances in embedding model distillation and cross-tokenizer preference alignment \cite{truong2025emo, an2026mol, truong2026ctpd} have demonstrated highly effective ways to compress and align these representation spaces. These techniques offer promising avenues for making cross-lingual topic alignment more computationally efficient and structurally robust in future iterations.

\textbf{MMD and Distribution Alignment.} Maximum Mean Discrepancy (MMD) formalizes distributional comparison by embedding probability measures into an RKHS and measuring distances between their kernel mean embeddings, a perspective that underlies modern two-sample and independence testing and reveals an equivalence to energy-distance statistics via negative-type semimetrics \cite{gretton2012kernel,muandet2017kernel,sejdinovic2013equivalence}. In practice, sensitivity to kernel scale makes bandwidth selection pivotal: the median heuristic is a strong default whose asymptotic behavior in testing settings has been clarified, while multi-kernel MMD (MK-MMD) mitigates scale mismatch by aggregating bandwidths and is widely adopted in deep representation learning \cite{garreau2017median,long2015dan}. For unsupervised domain adaptation, minimizing MMD between source and target features—popularized by Deep Adaptation Networks—yields effective feature alignment, and \emph{weighted} MMD further corrects for class-prior shift when label proportions differ across domains \cite{long2015dan,yan2017weighted}. Beyond testing and adaptation, MMD serves directly as a learning objective for generative modeling: Generative Moment Matching Networks match data and model distributions by minimizing MMD, and MMD-GAN variants learn kernels adversarially to improve sample fidelity, stability, and training efficiency \cite{li2015gmmn,li2017mmdgan}.

\section{Algorithm}
In this section, we present the \textbf{LLM-XTM} training procedure:
% Preamble cần có:
% \usepackage{algorithm}
% \usepackage{algorithmic}
% \usepackage{float} % nếu dùng [H]
\begin{algorithm}[t]
\caption{LLM-XTM training procedure}
\label{alg:llmxtm_training}
\begin{algorithmic}[1]
    \Require Input corpus $\mathbf{X}=\mathbf{X}^{(1)}\cup\mathbf{X}^{(2)}$, topic number $K$, top-$N$ words, refinement rounds $R$, refinement frequency $F$, temperature $\tau$, loss weights $\lambda_{\mathrm{mmd}},\lambda_{\mathrm{qa}}$.
    \Ensure Optimized parameters $\Theta^*=\{\text{Encoders}^*,\beta^{(1)*},\beta^{(2)*}\}$.
    \State Initialize pretrained backbone cross-lingual topic model parameters $\Theta$ and optimizer.
    \For{epoch $e=1$ to $N$}
        \State Train backbone to compute $\theta_d$ and $\beta^{(1)},\beta^{(2)}$ for mini-batches.
        \If{$e \bmod F = 0$} \Comment{Refinement step}
            \For{each topic $k=1$ to $K$}
                \State Extract top-$N$ words from $\beta^{(1)},\beta^{(2)}$.
                \State Run $R$ LLM refinements and vote to obtain refined set $\bar{w}_k$.
            \EndFor
            \State Build refined topic–word distributions $\beta^{(\mathrm{refined})}$.
        \EndIf
        \State Construct raw distributions $\beta^{(\mathrm{raw})}$.
        \State Compute $L_{\mathrm{MMD}}$ between $\beta^{(\mathrm{raw})}$ and $\beta^{(\mathrm{refined})}$.
        \State Encode documents $h_d$ and topics $t_k$ with a multilingual encoder.
        \State Derive target $\hat{\theta}_d=\mathrm{softmax}(\cos(h_d,t_k)/\tau)$.
        \State Compute $L_{\mathrm{doc}}=\sum_d \mathrm{KL}(\theta_d\parallel \hat{\theta}_d)$.
        \State Update $\Theta$ by minimizing $\mathcal{J}=L_{\text{Phase1}}+\lambda_{\mathrm{mmd}}L_{\mathrm{MMD}}+\lambda_{\mathrm{qa}}L_{\mathrm{doc}}$.
    \EndFor
\end{algorithmic}
\end{algorithm}

\section{Dataset Statistics}
\label{sec:dataset_statistics}
\begin{itemize}
    \item \textbf{EC News.} This corpus comprises parallel English–Chinese news articles covering six domains: business, education, entertainment, sports, technology, and fashion. Each language maintains a capped vocabulary of 5,000 tokens. The data are divided into 77{,}480 training samples (37{,}480 English, 40{,}000 Chinese) and 19{,}370 test samples (9{,}370 English, 10{,}000 Chinese).

    \item \textbf{Amazon Review.} This dataset contains bilingual product reviews in English and Chinese. We follow a binary sentiment setup, where reviews with a 5-star rating are labeled as positive and all others as negative. Both vocabularies are restricted to 5{,}000 words. The split includes 40{,}000 documents for training (20{,}000 per language) and 10{,}000 for evaluation (5{,}000 per language).

    \item \textbf{Rakuten Amazon.} This collection combines Japanese reviews from Rakuten with English reviews from Amazon, also used for binary sentiment prediction. Each language keeps a 5{,}000-word vocabulary limit, and the data are partitioned into 40{,}000 training and 10{,}000 test documents, balanced across languages.
\end{itemize}

For consistency with previous studies, we reproduce the experimental configuration of InfoCTM~\cite{infoctm} and employ its publicly released preprocessed datasets: \textit{EC News}, \textit{Amazon Review}, and \textit{Rakuten Amazon}.  

Unlike the original setup, both word- and document-level embeddings in our framework are derived from the \texttt{BAAI/bge-m3} model~\cite{bgem3}, ensuring a unified multilingual semantic space for cross-lingual topic alignment.

\begin{table*}[!htbp]
\centering
\small
% giảm khoảng cách cột để tiết kiệm chiều ngang
\setlength{\tabcolsep}{2.2mm}
\renewcommand{\arraystretch}{1.1}
% bó bảng theo đúng bề rộng hai cột
\resizebox{\textwidth}{!}{%
\begin{tabular}{l cccc cccc cccc}
\toprule
\multirow{2}{*}{\textbf{Model}} 
  & \multicolumn{4}{c}{\textbf{EC News}} 
  & \multicolumn{4}{c}{\shortstack{\textbf{Amazon Review}}} 
  & \multicolumn{4}{c}{\shortstack{\textbf{Rakuten Amazon}}} \\
\cmidrule(lr){2-5} \cmidrule(lr){6-9} \cmidrule(lr){10-13}
 & EN-I & EN-C & ZH-I & ZH-C & EN-I & EN-C & ZH-I & ZH-C & EN-I & EN-C & JA-I & JA-C \\
\midrule
InfoCTM 
& 0.7813 & 0.5242 & 0.7652 & 0.5592 & 0.7894 & 0.6862 & 0.7262 & 0.5984 & 0.7940 & 0.6910 & 0.8256 & 0.7342
\\
{+ LLM-XTM} 
& 0.7849 & 0.5445 & 0.7682 & 0.5875 & 0.8003 & 0.6863 & 0.7321 & 0.6043 & 0.7984 & 0.7314 & 0.8296 & 0.7875 \\
& $\uparrow$\,0.5\% & $\uparrow$\,3.9\% & $\uparrow$\,0.4\% & $\uparrow$\,5.1\%
& $\uparrow$\,1.4\% & $\uparrow$\,0.0\% & $\uparrow$\,0.8\% & $\uparrow$\,1.0\%
& $\uparrow$\,0.6\% & $\uparrow$\,5.8\% & $\uparrow$\,0.5\% & $\uparrow$\,7.3\% \\
\midrule
NMTM 
& 0.7917 & 0.4941 & 0.7742 & 0.5139 & 0.7859 & 0.5920 & 0.7210 & 0.5750 & 0.7966 & 0.6104 & 0.8264 & 0.6816
 \\
{+ LLM-XTM} 
& 0.7975 & 0.5173 & 0.7785 & 0.5333 & 0.7859 & 0.6513 & 0.7071 & 0.6085 & 0.7923 & 0.6211 & 0.8334 & 0.7282
\\
& $\uparrow$\,0.7\% & $\uparrow$\,4.7\% & $\uparrow$\,0.6\% & $\uparrow$\,3.8\%
& $\uparrow$\,0.0\% & $\uparrow$\,10.0\% & $\downarrow$\,1.9\% & $\uparrow$\,5.8\%
& $\downarrow$\,0.5\% & $\uparrow$\,1.8\% & $\uparrow$\,0.9\% & $\uparrow$\,6.8\% \\
\bottomrule
\end{tabular}%
}
\caption{Document classification accuracy using document--topic vectors ($\theta$) as features for linear SVMs. intra-lingual (-I) and cross-lingual (-C) accuracy across \textit{EC News}, \textit{Amazon Review}, and \textit{Rakuten Amazon}.}
\label{tab:classification_accuracy}
\end{table*}
\section{Classification Performance Within and Across Languages}
Following standard practice, we use document--topic representations as input features for linear SVMs and evaluate both intra-lingual (-I) and cross-lingual (-C) classification accuracy on \textit{EC News}, \textit{Amazon Review}, and \textit{Rakuten Amazon} (Table~\ref{tab:classification_accuracy}). Models without document–topic posteriors are excluded from comparison.

Across all datasets, integrating \textbf{LLM-XTM} consistently improves cross-lingual transfer while maintaining strong in-language performance. The largest gains appear on \textit{Rakuten Amazon} (–C: 0.734~$\rightarrow$~0.788 for InfoCTM; 0.682~$\rightarrow$~0.728 for NMTM), with smaller but stable improvements on \textit{EC News} and \textit{Amazon Review}. In-language accuracy remains comparable or slightly higher (e.g., InfoCTM EN-I: 0.789~$\rightarrow$~0.800).

These results show that \textbf{LLM-XTM} enhances cross-lingual generalization without compromising monolingual accuracy, consistent with its design objectives of MMD-based refinement and QA-driven alignment.
\section{Prompt for Cross-Lingual Topic Refinement}
\label{app:prompt}

We provide the exact prompt used to query the Gemini API for topic refinement. 
The same template is applied to all topics, with the top-15 words from each 
language filled in.

\begin{figure}[H]
\centering
\begin{minipage}{0.95\linewidth}
\ttfamily\small
\begin{flushleft}
Given the following cross-lingual topic words from English and Chinese for N topics, refine each topic:

1) Identify the main theme shared across both languages.\\
2) Remove irrelevant/noisy words that do not fit the theme.\\
  3) Add relevant words that strengthen coherence and cross-lingual coverage.\\
4) Use only SINGLE WORDS (no phrases, no underscores, no hyphenated expressions).\\
5) Return exactly 15 words per language for each topic.

\textbf{Output format for all topics:}\\
Topic <id>: <brief theme>\\
EN: word1 - word2 - ... - word15\\
CN: word1 - word2 - ... - word15

\textbf{Rules:}\\
- Exactly 15 words after EN: and CN:.\\
- Separate words with " - ".\\
- List topics in order from 0 to N–1.
\end{flushleft}
\end{minipage}
\caption{Prompt used for cross-lingual topic refinement}
\end{figure}

\section{Detailed Prompts for LLM Evaluation} % Tiêu đề cho phần phụ lục này
\label{sec:llm_prompts} % Nhãn để tham chiếu từ thân bài

This appendix provides the detailed system prompts used for the LLM-based evaluation tasks described in the main text. Tables~\ref{tab:intra-lingual_prompts} and \ref{tab:cross-lingual_prompts} shows the prompts side-by-side for intra-lingual coherence and cross-lingual similarity assessment across the different datasets.

\begin{table*}[htbp]
\centering
\begin{tabular}{|>{\centering\arraybackslash}p{0.1\textwidth}|p{0.85\textwidth}|}
\hline
\rowcolor{gray!10}
\textbf{Dataset} & \multicolumn{1}{c|}{\textbf{Prompt}} \\
\hline

\multirow{6}{*}{\parbox[c]{0.1\textwidth}{\centering\textbf{EC\\[0.5ex] News}}} & 
You are a helpful assistant evaluating the top words of a topic model output for a given topic. The dataset is EC News, a collection of English and Chinese news with 6 categories: business, education, entertainment, sports, tech, and fashion. Please rate how related the following words are to each other on a scale from 1 to 3 ("1"=not very related, "2"=moderately related, "3"=very related). Reply with a single number, indicating the overall appropriateness of the topic. \\
\hline

\multirow{7}{*}{\parbox[c]{0.1\textwidth}{\centering\textbf{Amazon Review}}} & 
You are a helpful assistant evaluating the top words of a topic model output for a given topic. The dataset is Amazon Review, which includes English and Chinese reviews from the Amazon website. Please rate how related the following words are to each other on a scale from 1 to 3 ("1"=not very related, "2"=moderately related, "3"=very related). Reply with a single number, indicating the overall appropriateness of the topic. \\
\hline

\multirow{6}{*}{\parbox[c]{0.1\textwidth}{\centering\textbf{Rakuten Amazon}}} & 
You are a helpful assistant evaluating the top words of a topic model output for a given topic. The dataset is Rakuten Amazon, which contains Japanese reviews from Rakuten, and English reviews from Amazon. Please rate how related the following words are to each other on a scale from 1 to 3 ("1"=not very related, "2"=moderately related, "3"=very related). Reply with a single number, indicating the overall appropriateness of the topic. \\
\hline

\end{tabular}
\caption{intra-lingual Coherence Prompts for LLM-based Evaluation}
\label{tab:intra-lingual_prompts}

\end{table*}

\begin{table*}[htbp]
\centering

\begin{tabular}{|>{\centering\arraybackslash}p{0.1\textwidth}|p{0.85\textwidth}|}
\hline
\rowcolor{gray!10}
\textbf{Dataset} & \multicolumn{1}{c|}{\textbf{Prompt}} \\
\hline

\multirow{6}{*}{\parbox[c]{0.1\textwidth}{\centering\textbf{EC\\[0.5ex] News}}} & 
You are a helpful assistant evaluating the similarity of topics derived from topic modeling on parallel news corpora. The dataset is EC News, with English and Chinese news. You will be given two sets of top words, one for an English topic (Language 1) and one for a Chinese topic (Language 2). Please rate how similar the underlying topics represented by these two sets of words are, on a scale from 1 to 3 ("1"=not very similar, "2"=moderately similar, "3"=very similar). Reply with a single number. \\
\hline

\multirow{6}{*}{\parbox[c]{0.1\textwidth}{\centering\textbf{Amazon Review}}} & 
You are a helpful assistant evaluating the similarity of topics derived from topic modeling on parallel review corpora. The dataset is Amazon Review, with English and Chinese reviews. You will be given two sets of top words, one for an English topic (Language 1) and one for a Chinese topic (Language 2). Please rate how similar the underlying topics represented by these two sets of words are, on a scale from 1 to 3 ("1"=not very similar, "2"=moderately similar, "3"=very similar). Reply with a single number. \\
\hline

\multirow{7}{*}{\parbox[c]{0.1\textwidth}{\centering\textbf{Rakuten Amazon}}} & 
You are a helpful assistant evaluating the similarity of topics derived from topic modeling on parallel review corpora. The dataset is Rakuten Amazon, with Japanese reviews (Rakuten - Language 2) and English reviews (Amazon - Language 1). You will be given two sets of top words, one for an English topic and one for a Japanese topic. Please rate how similar the underlying topics represented by these two sets of words are, on a scale from 1 to 3 ("1"=not very similar, "2"=moderately similar, "3"=very similar). Reply with a single number. \\
\hline

\end{tabular}
\caption{cross-lingual Similarity Prompts for LLM-based Evaluation}
\label{tab:cross-lingual_prompts}

\end{table*}

\section{Implementation Details}

We run experiments on a single NVIDIA P100 GPU (Kaggle). Training is divided into two phases: a \textbf{base training} phase , followed by \textbf{LLM-based refinement} phase of 30 epochs.

Refinement is performed over 5 rounds using these settings:
\begin{itemize}
  \item Refinement frequency: \{8, 10\} steps  
  \item MMD loss weight: 20\,000  
  \item Document alignment loss weight: \{100, 200, 300\}  
\end{itemize}
\section{Generalization to Long Documents (Airiti Thesis)}
\label{sec:airiti_appendix}

To test generalization beyond short reviews and news, we evaluate LLM-XTM on the Airiti Thesis dataset introduced by Chang et al.~\cite{refining}, which contains longer academic thesis abstracts in Chinese and English. Following \citet{refining}, we use the same data source and preprocessing setting; all CNPMI scores in Table~\ref{tab:airiti_results} are computed against the Wikipedia reference corpus, so coherence is measured independently of the training text.

\begin{table}[t]
\centering
\small
\setlength{\tabcolsep}{5pt}
\renewcommand{\arraystretch}{1.08}
\begin{tabular}{llccc}
\toprule
\textbf{Model} & \textbf{Configuration} & \textbf{CNPMI} & \textbf{TU} & \textbf{TQ} \\
\midrule
u-SVD   & Baseline     & 0.0281 & 0.5893 & 0.0165 \\
SVD-LR  & Baseline     & 0.0312 & 0.5940 & 0.0185 \\
\midrule
InfoCTM & Base         & 0.0279 & 0.7953 & 0.0222 \\
        & + LLM-XTM    & 0.0497 & 0.7213 & 0.0358 \\
        & Improvement  & +78.1\% & -9.3\% & +61.3\% \\
\midrule
XTRA    & Base         & 0.0106 & 0.7087 & 0.0075 \\
        & + LLM-XTM    & 0.0207 & \textbf{0.8027} & 0.0166 \\
        & Improvement  & +95.3\% & +13.3\% & +121.3\% \\
\midrule
NMTM    & Base         & 0.0417 & 0.6213 & 0.0259 \\
        & + LLM-XTM    & \textbf{0.0531} & 0.6987 & \textbf{0.0371} \\
        & Improvement  & +27.3\% & +12.6\% & +43.2\% \\
\bottomrule
\end{tabular}
\caption{Results on the Airiti Thesis long-document benchmark~\cite{refining}. CNPMI is computed using the Wikipedia reference corpus. Bold indicates the best value in each metric column among the reported systems.}
\label{tab:airiti_results}
\end{table}

\begin{table*}[t]
\centering
\small
\setlength{\tabcolsep}{5pt}
\renewcommand{\arraystretch}{1.08}
\begin{tabular}{lccccc}
\toprule
\textbf{Refinement Engine} & \textbf{Model Size} & \textbf{CNPMI} & \textbf{TU} & \textbf{TQ} & \textbf{vs. Baseline} \\
\midrule
Base Model (NMTM)                  & --               & 0.0430 & 0.6100 & 0.0262 & -- \\
Gemini 2.5 Flash                   & --               & \textbf{0.0560} & 0.6270 & \textbf{0.0351} & +34.0\% \\
Llama-3.3-70B                      & 70B              & 0.0544 & 0.6453 & \textbf{0.0351} & +34.0\% \\
Qwen3-Coder-480B-A35B-Instruct     & 480B (35B active) & 0.0508 & 0.6567 & 0.0334 & +27.5\% \\
Mistral-Small-24B                  & 24B              & 0.0506 & 0.6413 & 0.0325 & +24.0\% \\
Mistral-7B-v0.3                    & 7B               & 0.0445 & \textbf{0.6673} & 0.0297 & +13.4\% \\
\bottomrule
\end{tabular}
\caption{Open-weight LLM comparison on Amazon Review with the NMTM backbone. Bold indicates the best value in each metric column.}
\label{tab:openweight_comparison}
\end{table*}
LLM-XTM yields substantial gains across all three neural backbones on this long-document benchmark. XTRA more than doubles Topic Quality (+121.3\%), InfoCTM improves by +61.3\%, and NMTM improves by +43.2\%. The strongest overall TQ is achieved by \textbf{NMTM + LLM-XTM} (0.0371), which is more than $2\times$ the best clustering baseline SVD-LR (0.0185), while Topic Uniqueness remains robust and even increases for both XTRA and NMTM.

\section{Model-Agnostic Evaluation with Open-Weight LLMs}
\label{sec:openweight_appendix}

To test whether the enhancement depends on a proprietary API, we replace Gemini 2.5 Flash with several open-weight alternatives on Amazon Review (NMTM backbone): Llama-3.3-70B~\cite{grattafiori2024llama3}, Qwen3-Coder-480B-A35B-Instruct~\cite{qwen3coder2025,qwen3report2025}, and Mistral-Small-24B~\cite{mistralsmall3}. We additionally include a compact 7B Mistral variant to probe the low-cost end of the trade-off. The quantitative comparison is reported in Table~\ref{tab:openweight_comparison}.

The results confirm that LLM-XTM is model-agnostic rather than tied to a single proprietary engine. As shown in Table~\ref{tab:openweight_comparison}, \textbf{Llama-3.3-70B} matches Gemini 2.5 Flash at TQ 0.0351, showing that an open-weight model can reach the same overall quality. \textbf{Qwen3-Coder-480B-A35B-Instruct} and \textbf{Mistral-Small-24B} still deliver strong gains (+27.5\% and +24.0\%), while even the 7B Mistral variant improves TQ by +13.4\%. The overall pattern suggests a predictable trade-off: smaller models preserve slightly higher uniqueness, whereas larger models maximize CNPMI and TQ.

\section{Additional LLM-based Evaluation Results}
\label{sec:llm_results}

% =======================
% INFOCTM
% =======================
\begin{figure}[H]
    \centering
    \includegraphics[width=\linewidth]{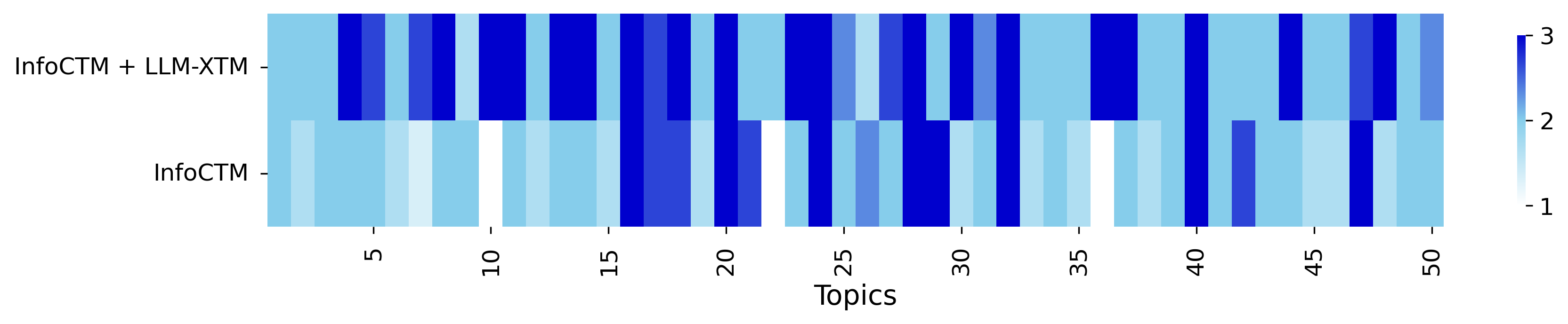}
    \caption{English intra-lingual semantic similarity (Amazon Review, InfoCTM).}
\end{figure}

\begin{figure}[H]
    \centering
    \includegraphics[width=\linewidth]{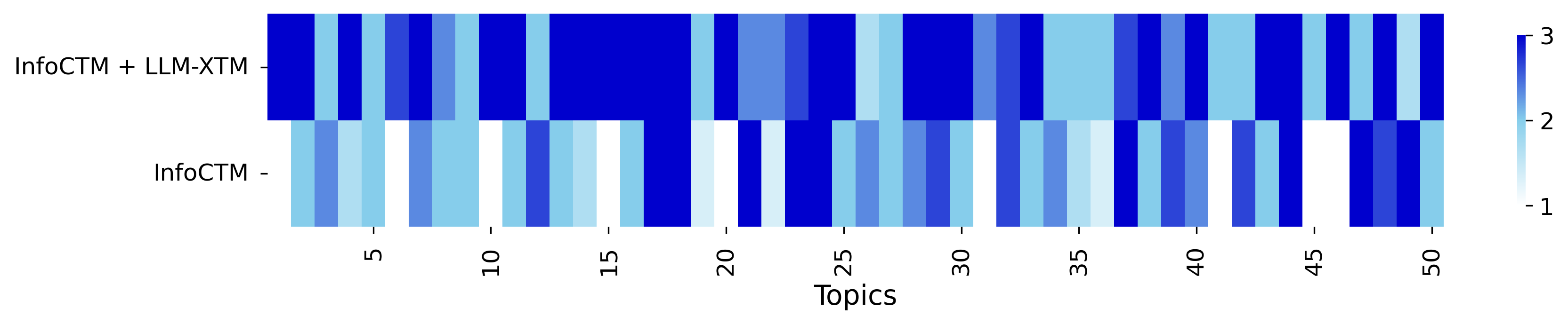}
    \caption{Chinese intra-lingual semantic similarity (Amazon Review, InfoCTM).}
\end{figure}

\begin{figure}[H]
    \centering
    \includegraphics[width=\linewidth]{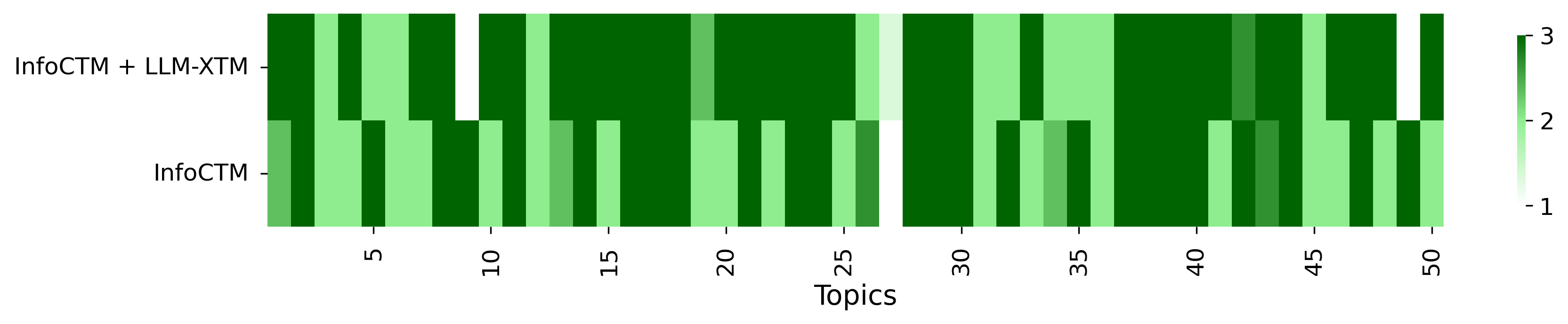}
    \caption{Cross-lingual semantic similarity on Amazon Review (InfoCTM).}
\end{figure}

\begin{figure}[H]
    \centering
    \includegraphics[width=\linewidth]{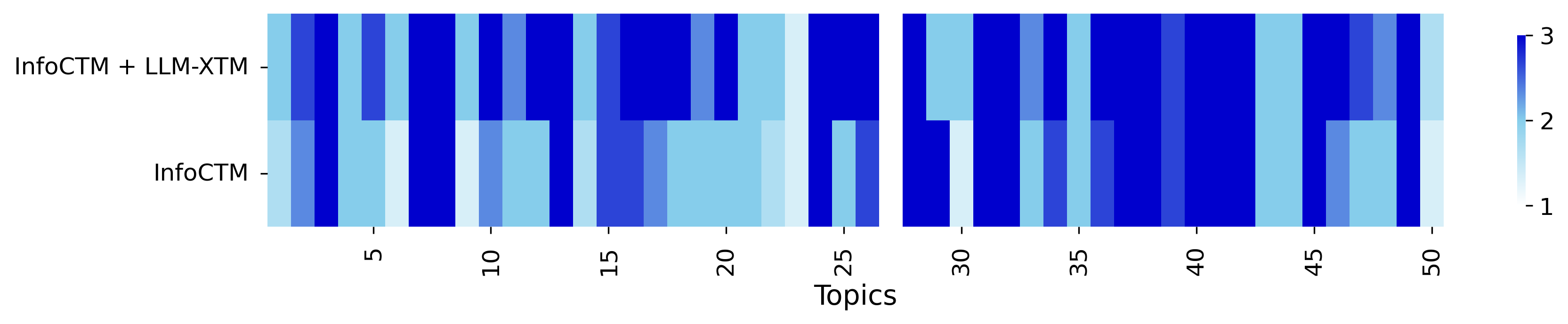}
    \caption{English intra-lingual semantic similarity (ECNews, InfoCTM).}
\end{figure}

\begin{figure}[H]
    \centering
    \includegraphics[width=\linewidth]{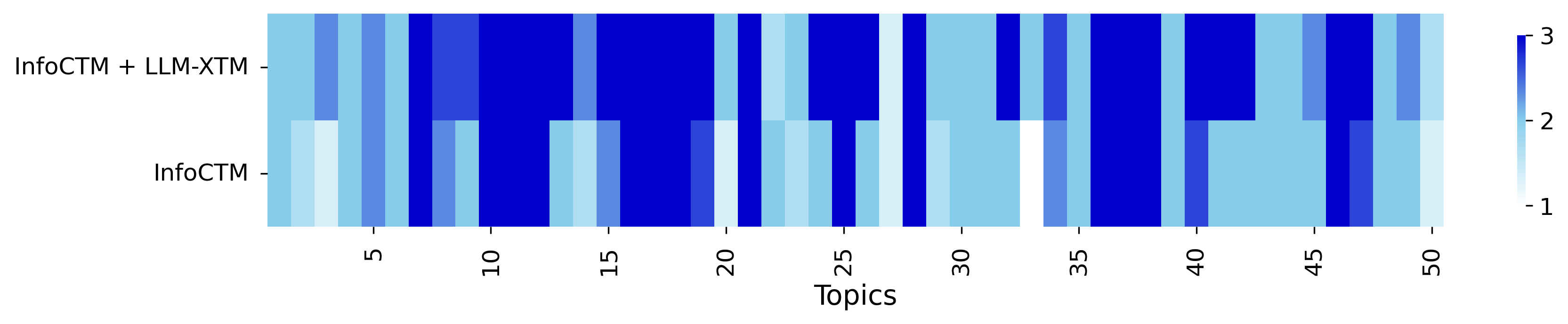}
    \caption{Chinese intra-lingual semantic similarity (ECNews, InfoCTM).}
\end{figure}

\begin{figure}[H]
    \centering
    \includegraphics[width=\linewidth]{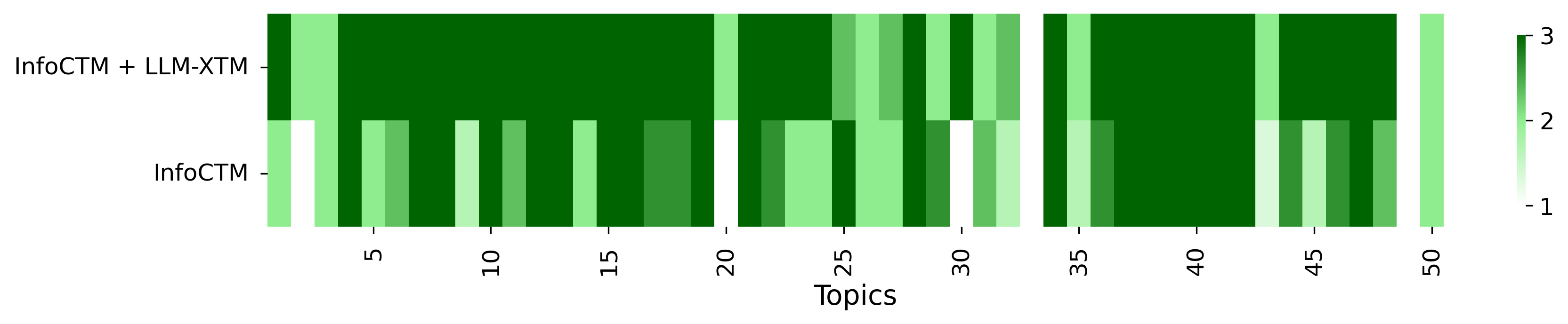}
    \caption{Cross-lingual semantic similarity on ECNews (InfoCTM).}
\end{figure}

\begin{figure}[H]
    \centering
    \includegraphics[width=\linewidth]{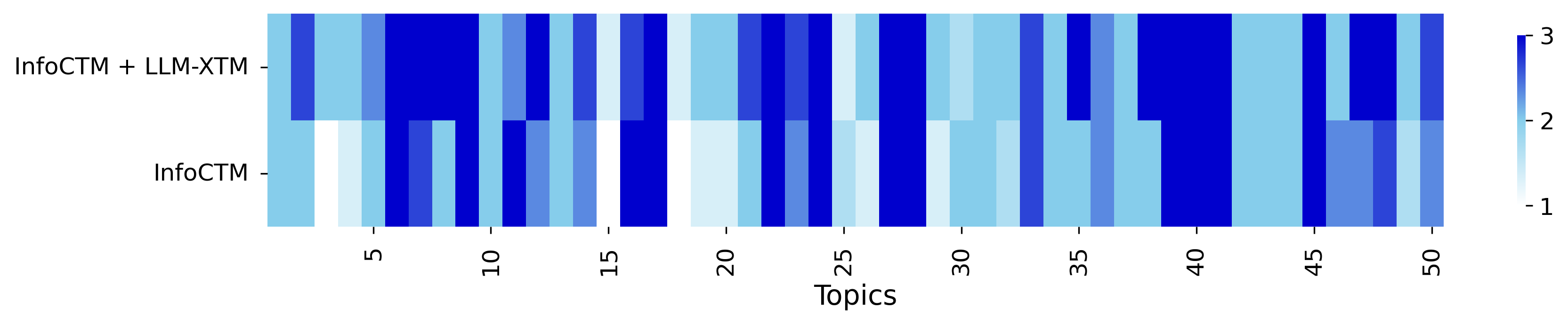}
    \caption{English intra-lingual semantic similarity (Rakuten\_Amazon, InfoCTM).}
\end{figure}

\begin{figure}[H]
    \centering
    \includegraphics[width=\linewidth]{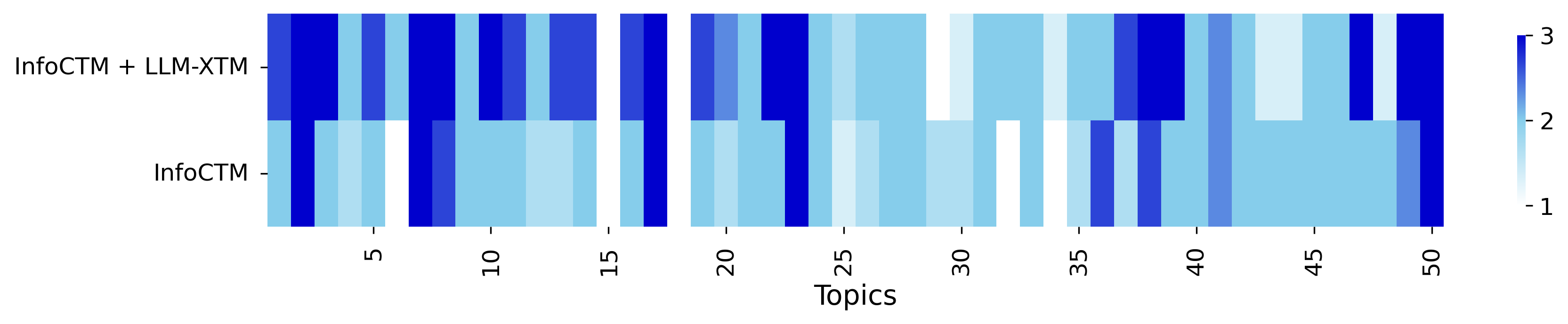}
    \caption{Japanese intra-lingual semantic similarity (Rakuten\_Amazon, InfoCTM).}
\end{figure}

\begin{figure}[H]
    \centering
    \includegraphics[width=\linewidth]{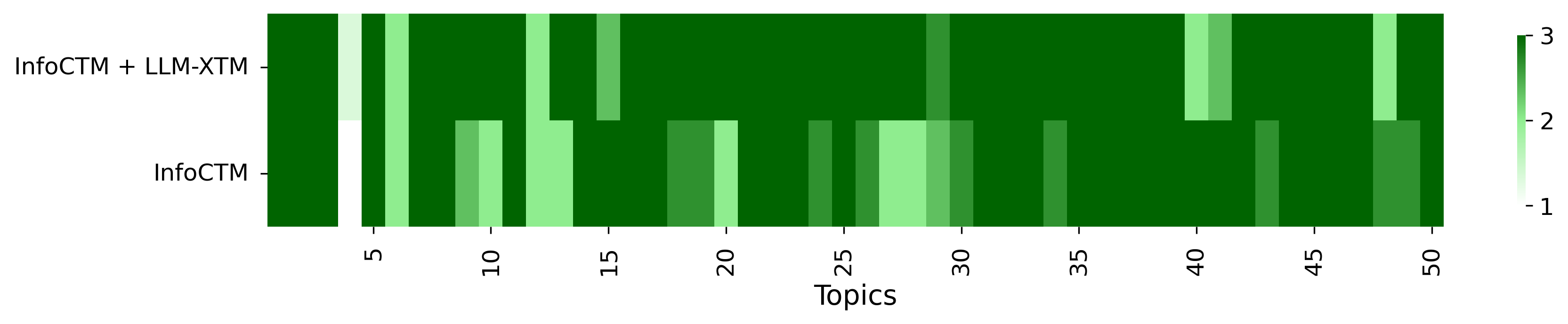}
    \caption{Cross-lingual semantic similarity on Rakuten\_Amazon (InfoCTM).}
\end{figure}

% =======================
% NMTM
% =======================
\begin{figure}[H]
    \centering
    \includegraphics[width=\linewidth]{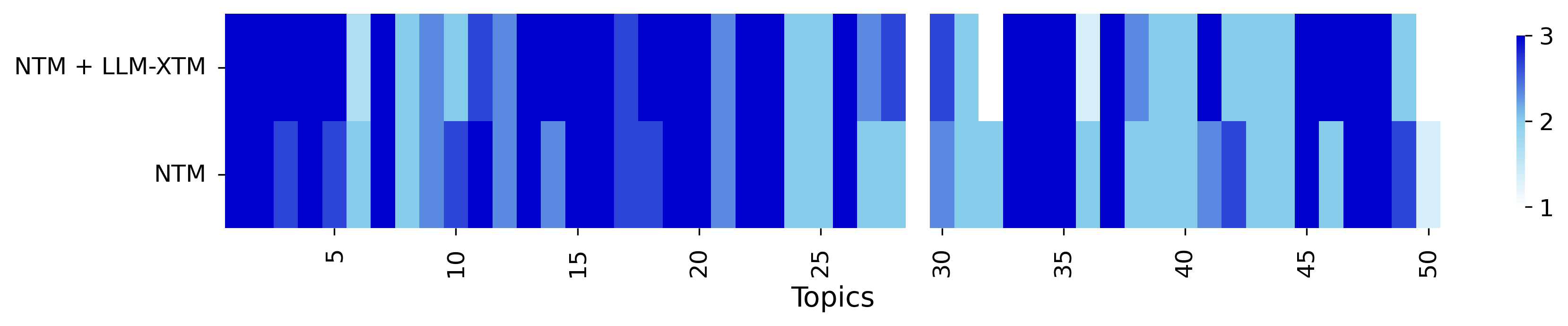}
    \caption{English intra-lingual semantic similarity (Amazon Review, NMTM).}
\end{figure}

\begin{figure}[H]
    \centering
    \includegraphics[width=\linewidth]{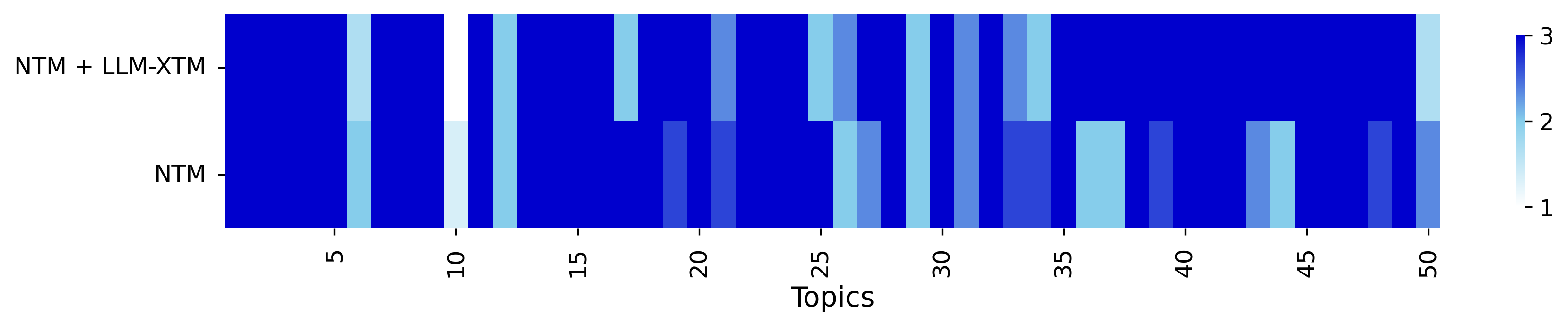}
    \caption{Chinese intra-lingual semantic similarity (Amazon Review, NMTM).}
\end{figure}

\begin{figure}[H]
    \centering
    \includegraphics[width=\linewidth]{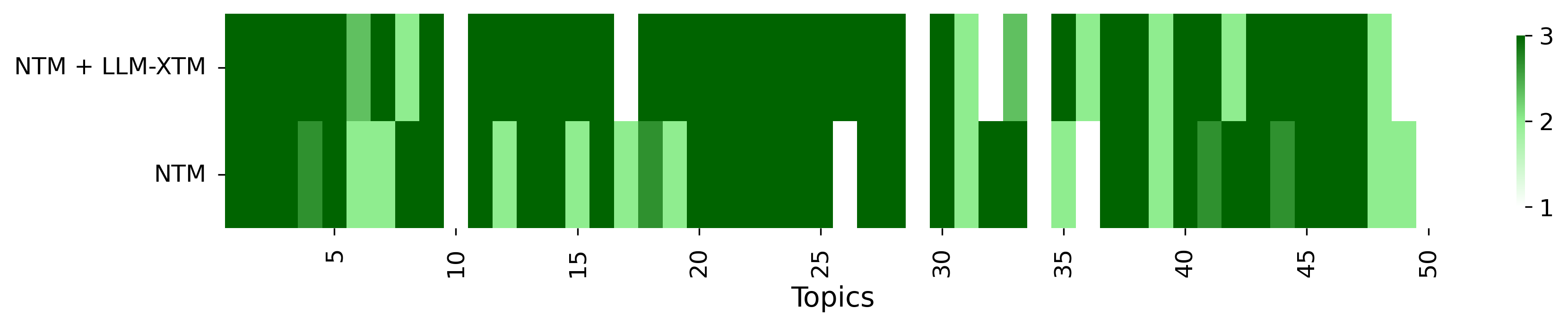}
    \caption{Cross-lingual semantic similarity on Amazon Review (NMTM).}
\end{figure}

\begin{figure}[H]
    \centering
    \includegraphics[width=\linewidth]{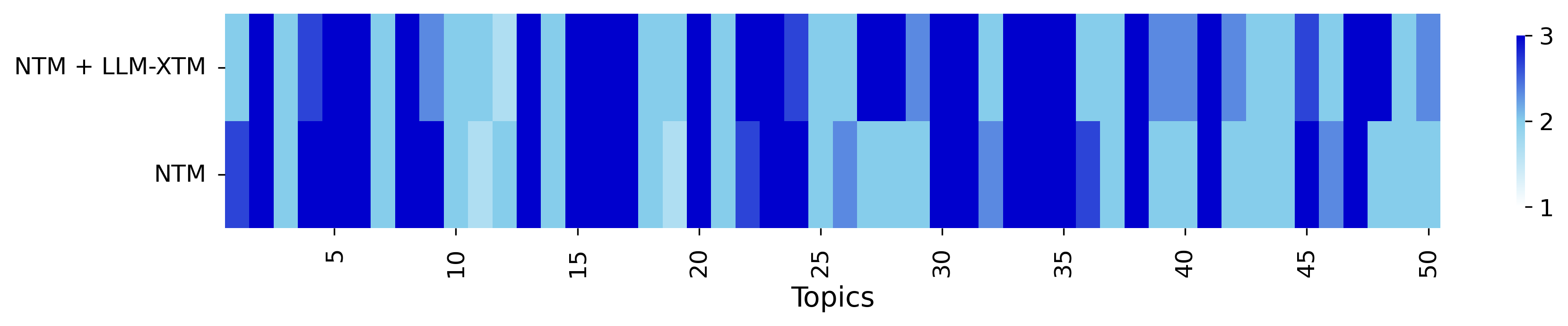}
    \caption{English intra-lingual semantic similarity (ECNews, NMTM).}
\end{figure}

\begin{figure}[H]
    \centering
    \includegraphics[width=\linewidth]{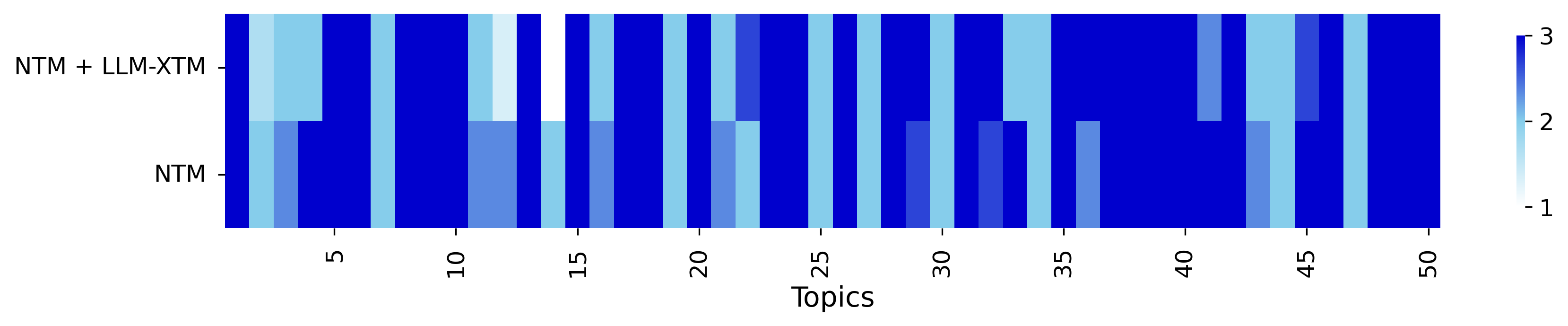}
    \caption{Chinese intra-lingual semantic similarity (ECNews, NMTM).}
\end{figure}

\begin{figure}[H]
    \centering
    \includegraphics[width=\linewidth]{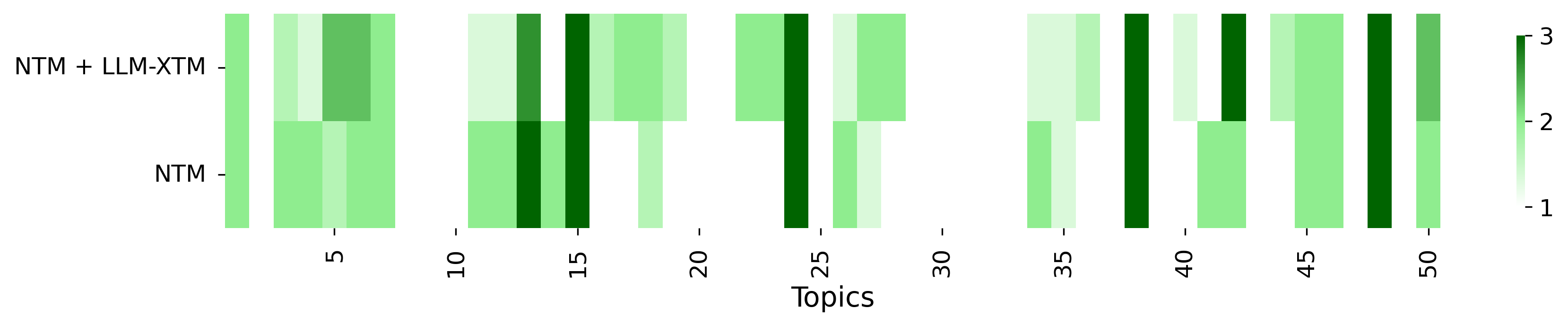}
    \caption{Cross-lingual semantic similarity on ECNews (NMTM).}
\end{figure}

\begin{figure}[H]
    \centering
    \includegraphics[width=\linewidth]{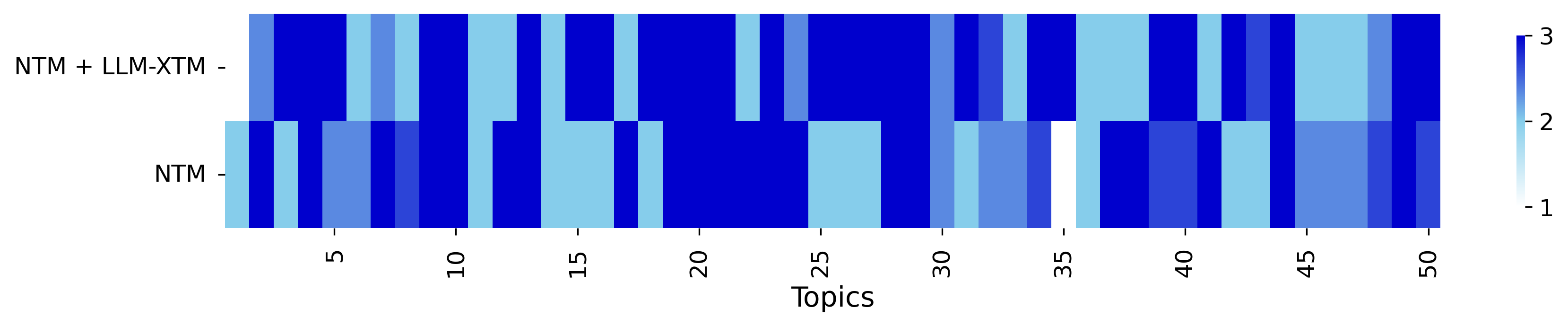}
    \caption{English intra-lingual semantic similarity (Rakuten\_Amazon, NMTM).}
\end{figure}

\begin{figure}[H]
    \centering
    \includegraphics[width=\linewidth]{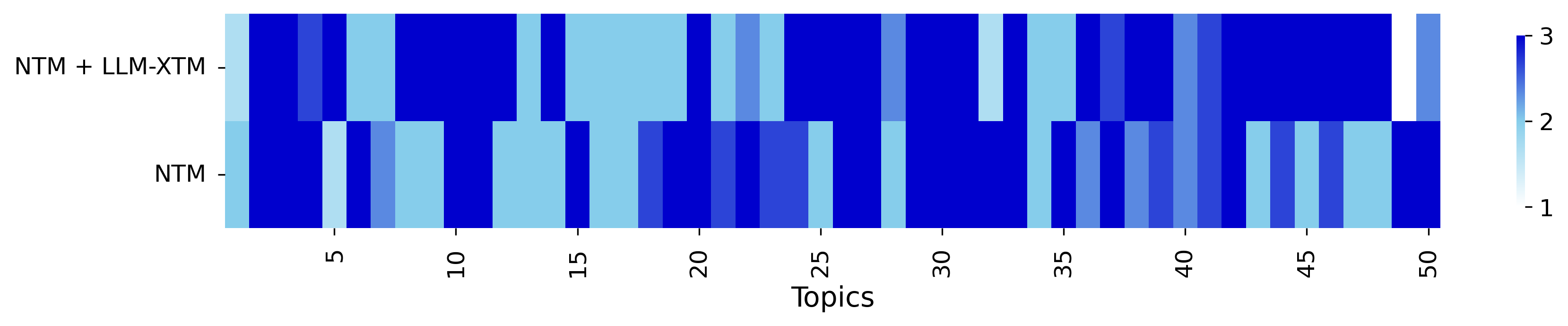}
    \caption{Japanese intra-lingual semantic similarity (Rakuten\_Amazon, NMTM).}
\end{figure}

\begin{figure}[H]
    \centering
    \includegraphics[width=\linewidth]{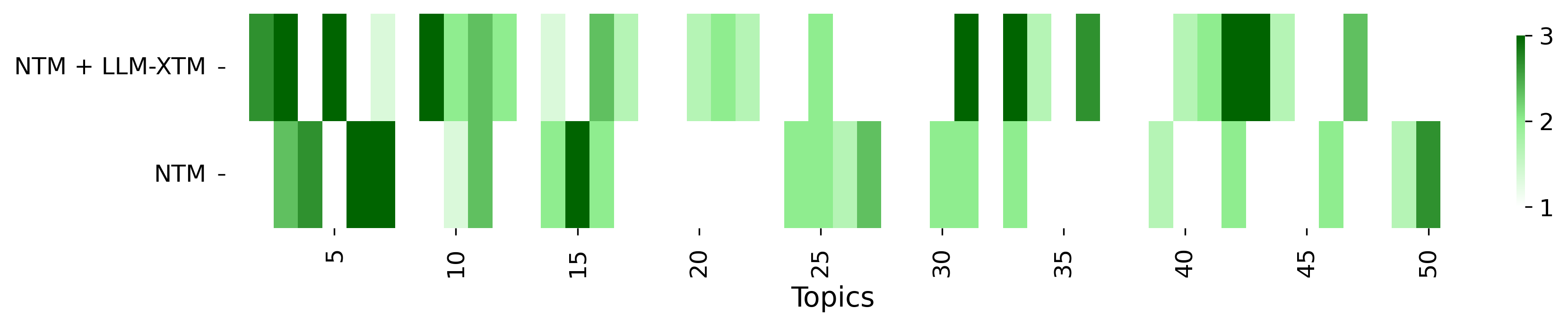}
    \caption{Cross-lingual semantic similarity on Rakuten\_Amazon (NMTM).}
\end{figure}

% =======================
% XTRA
% =======================
\begin{figure}[H]
    \centering
    \includegraphics[width=\linewidth]{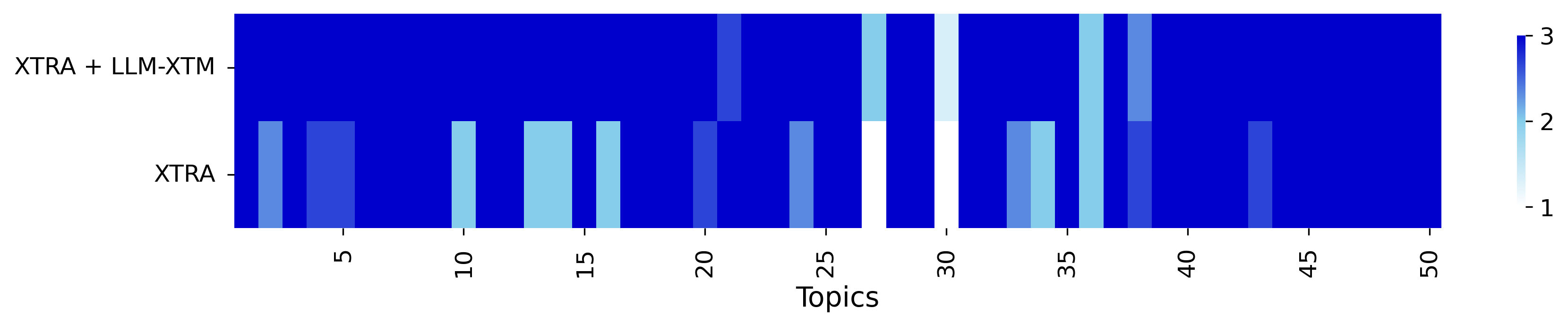}
    \caption{English intra-lingual semantic similarity (Amazon Review, XTRA).}
\end{figure}

\begin{figure}[H]
    \centering
    \includegraphics[width=\linewidth]{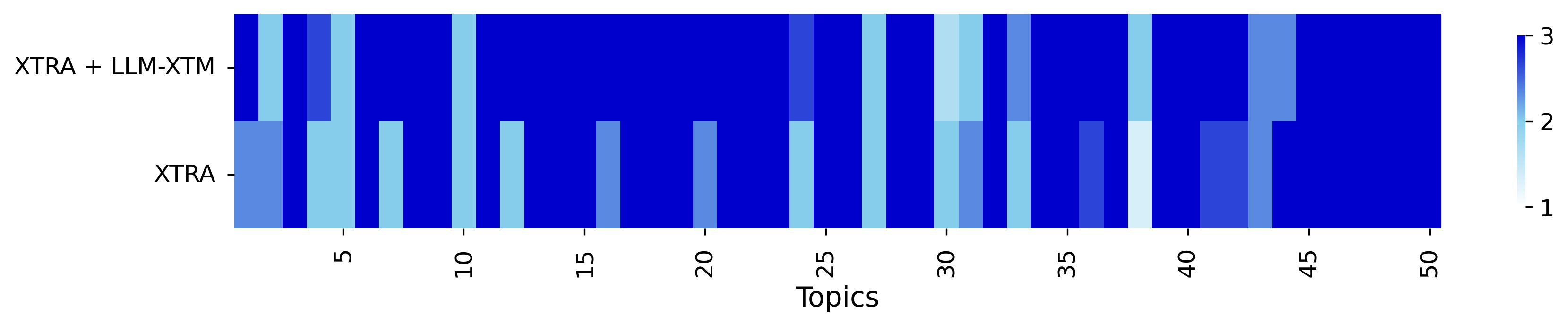}
    \caption{Chinese intra-lingual semantic similarity (Amazon Review, XTRA).}
\end{figure}

\begin{figure}[H]
    \centering
    \includegraphics[width=\linewidth]{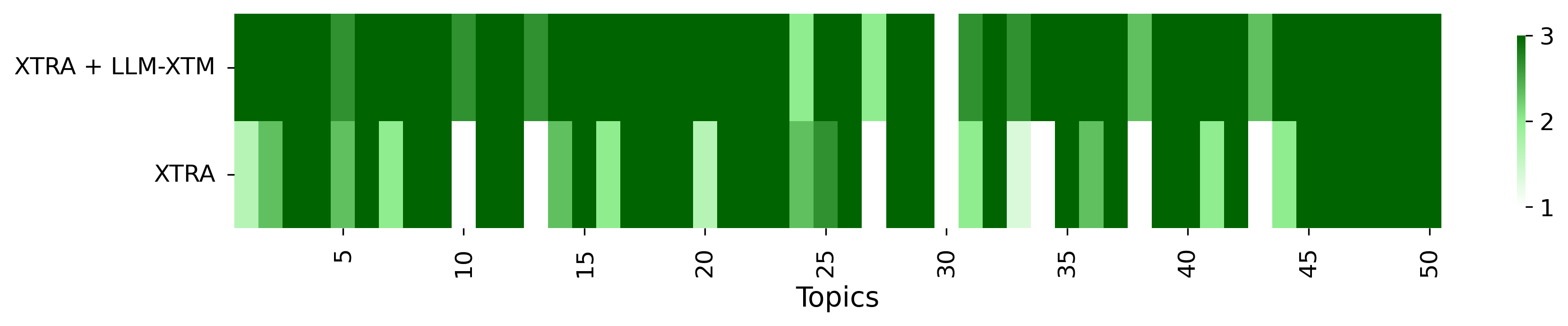}
    \caption{Cross-lingual semantic similarity on Amazon Review (XTRA).}
\end{figure}

\begin{figure}[H]
    \centering
    \includegraphics[width=\linewidth]{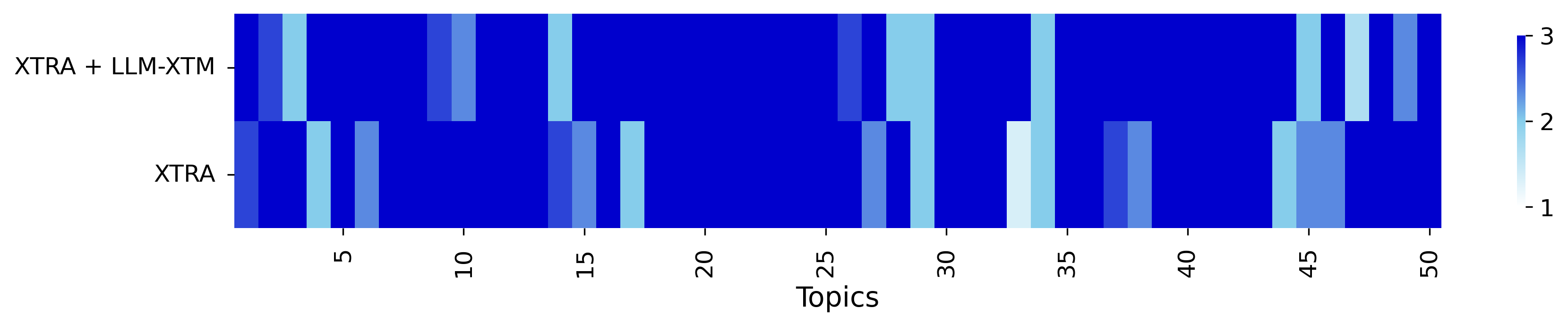}
    \caption{English intra-lingual semantic similarity (ECNews, XTRA).}
\end{figure}

\begin{figure}[H]
    \centering
    \includegraphics[width=\linewidth]{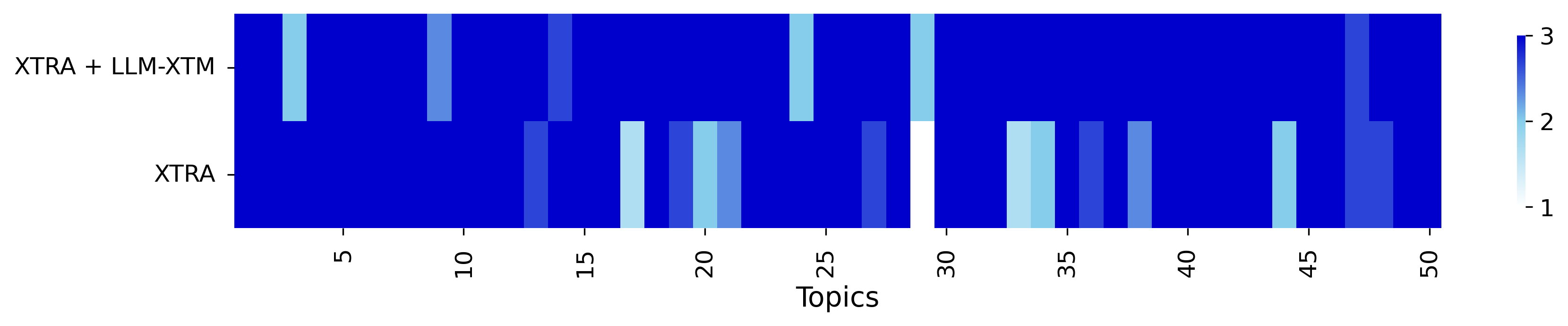}
    \caption{Chinese intra-lingual semantic similarity (ECNews, XTRA).}
\end{figure}

\begin{figure}[H]
    \centering
    \includegraphics[width=\linewidth]{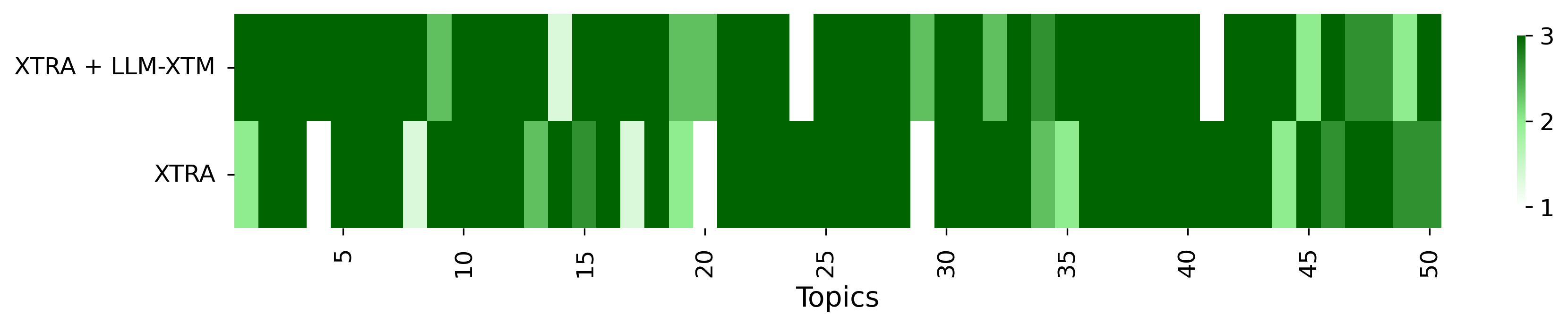}
    \caption{Cross-lingual semantic similarity on ECNews (XTRA).}
\end{figure}

\begin{figure}[H]
    \centering
    \includegraphics[width=\linewidth]{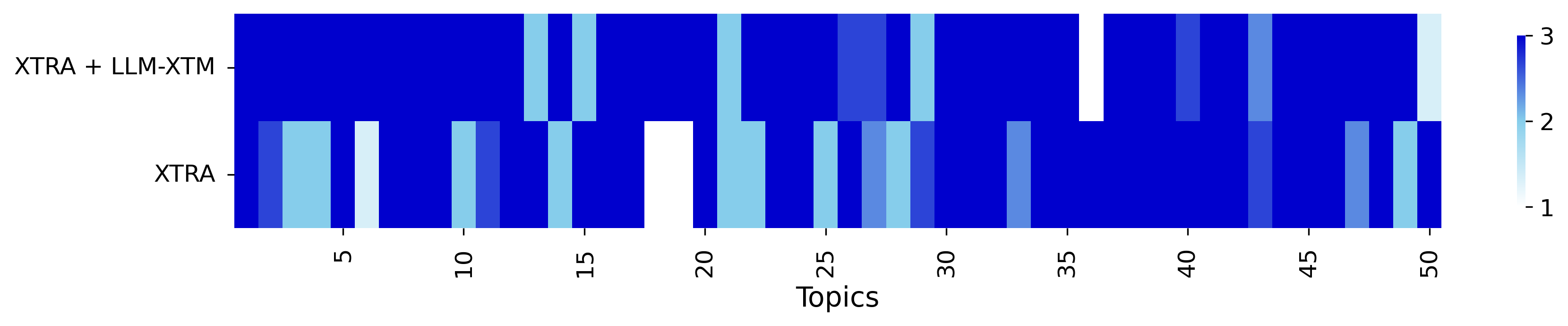}
    \caption{English intra-lingual semantic similarity (Rakuten\_Amazon, XTRA).}
\end{figure}

\begin{figure}[H]
    \centering
    \includegraphics[width=\linewidth]{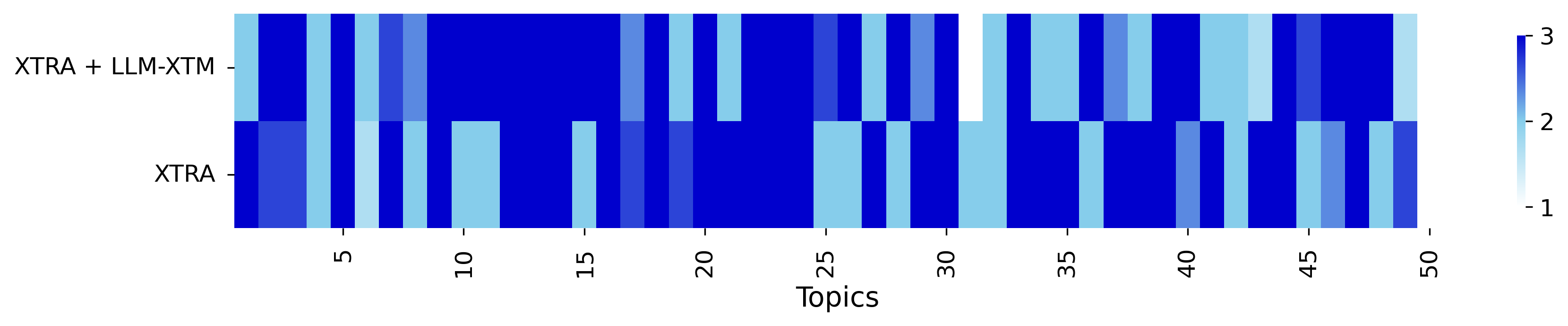}
    \caption{Japanese intra-lingual semantic similarity (Rakuten\_Amazon, XTRA).}
\end{figure}

\begin{figure}[H]
    \centering
    \includegraphics[width=\linewidth]{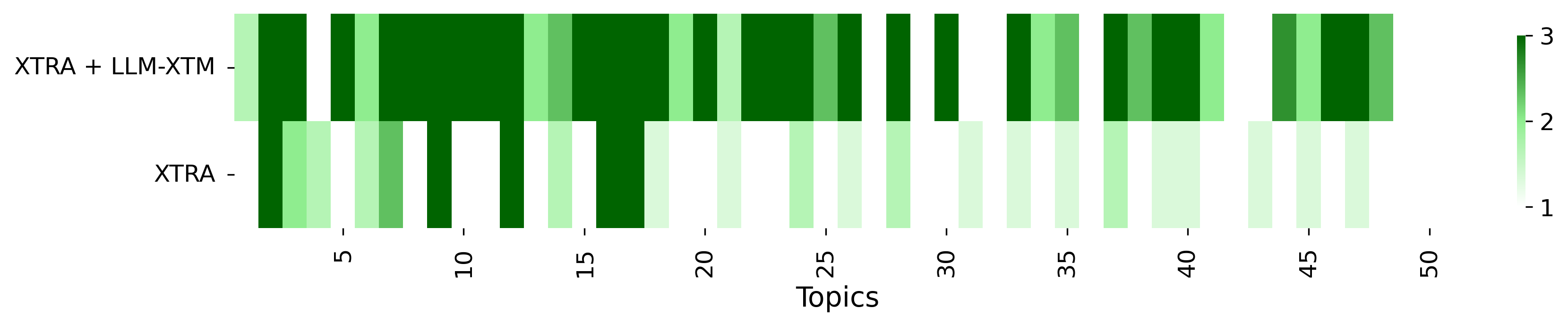}
    \caption{Cross-lingual semantic similarity on Rakuten\_Amazon (XTRA).}
\end{figure}

\end{document}